\documentclass[10pt,twocolumn,letterpaper]{article}

\usepackage{cvpr}
\usepackage{times}
\usepackage{epsfig}
\usepackage{graphicx}
\usepackage{amsmath}
\usepackage{amssymb}
\usepackage{subfigure}
\usepackage[table]{xcolor}

\usepackage[pagebackref=true,breaklinks=true,letterpaper=true,colorlinks,bookmarks=false]{hyperref}

 \cvprfinalcopy 

\def\cvprPaperID{441} 

\ifcvprfinal\fi
\begin{document}

\title{Transferring Rich Feature Hierarchies for Robust Visual Tracking}

\author{{Naiyan Wang}$^\dagger$ \quad {Siyi Li}$^\dagger$ \quad {Abhinav Gupta}$^\ddagger$ \quad {Dit-Yan Yeung}$^\dagger$ \\
	$^\dagger$ Hong Kong University of Science and Technology \quad \quad 
	$^\ddagger$ Carnegie Mellon University \\
	{\tt\small winsty@gmail.com} \quad {\tt\small sliay@cse.ust.hk} \quad {\tt \small abhinavg@cs.cmu.edu} \quad {\tt\small dyyeung@cse.ust.hk}
} 

\maketitle

\begin{abstract}
Convolutional neural network (CNN) models have demonstrated great success in various computer vision tasks including image classification and object detection.  However, some equally important tasks such as visual tracking remain relatively unexplored.  We believe that a major hurdle that hinders the application of CNN to visual tracking is the lack of properly labeled training data.  While existing applications that liberate the power of CNN often need an enormous amount of training data in the order of millions, visual tracking applications typically have only one labeled example in the first frame of each video.  We address this research issue here by pre-training a CNN offline and then transferring the rich feature hierarchies learned to online tracking.  The CNN is also fine-tuned during online tracking to adapt to the appearance of the tracked target specified in the first video frame. To fit the characteristics of object tracking, we first pre-train the CNN to recognize what is an object, and then propose to generate a probability map instead of producing a simple class label. Using two challenging open benchmarks for performance evaluation, our proposed tracker has demonstrated substantial improvement over other state-of-the-art trackers.
\end{abstract}

\section{Introduction}

\begin{figure}[t]
	\center{
	\includegraphics[width=0.32\linewidth]{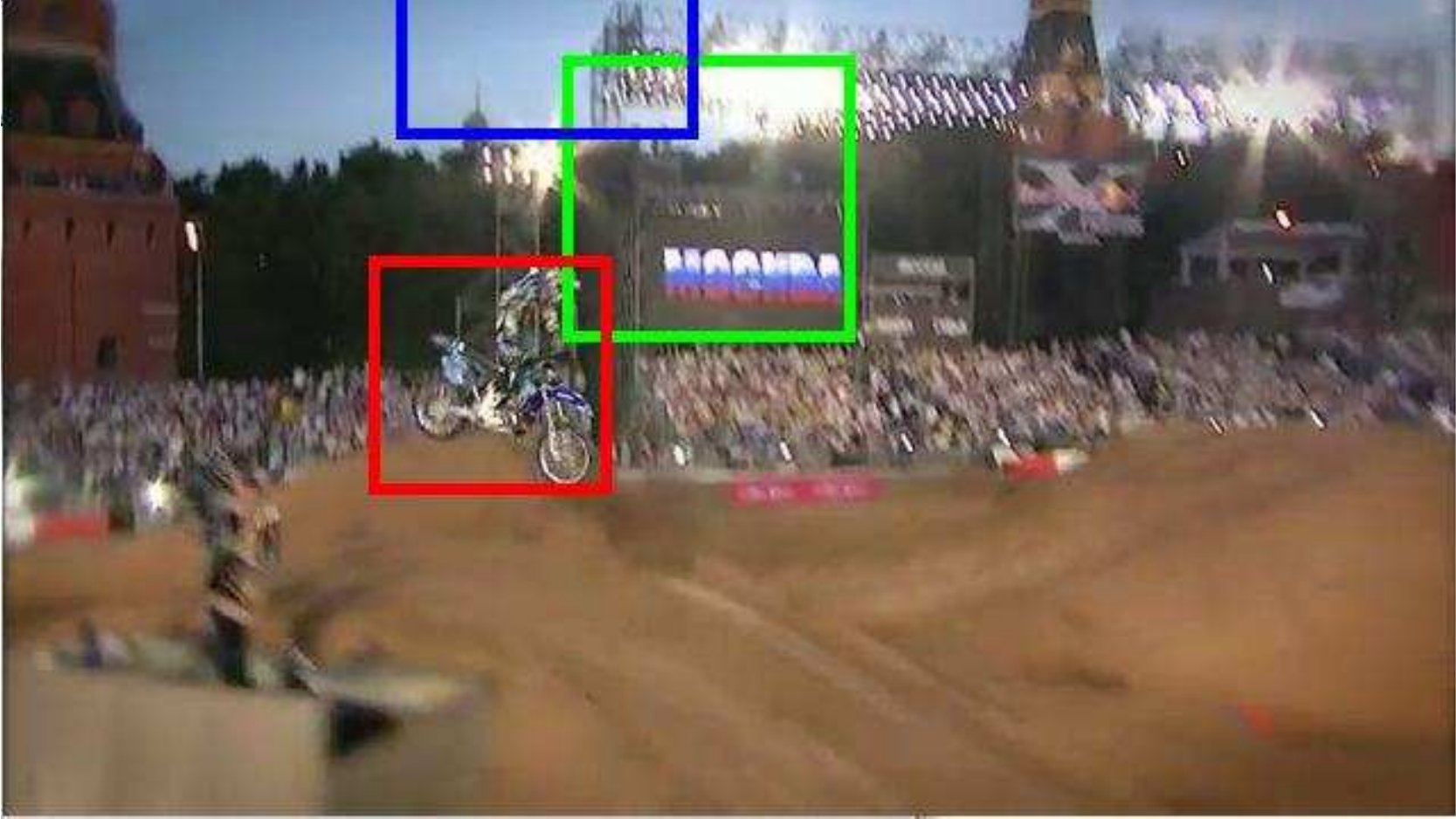}
	\includegraphics[width=0.32\linewidth]{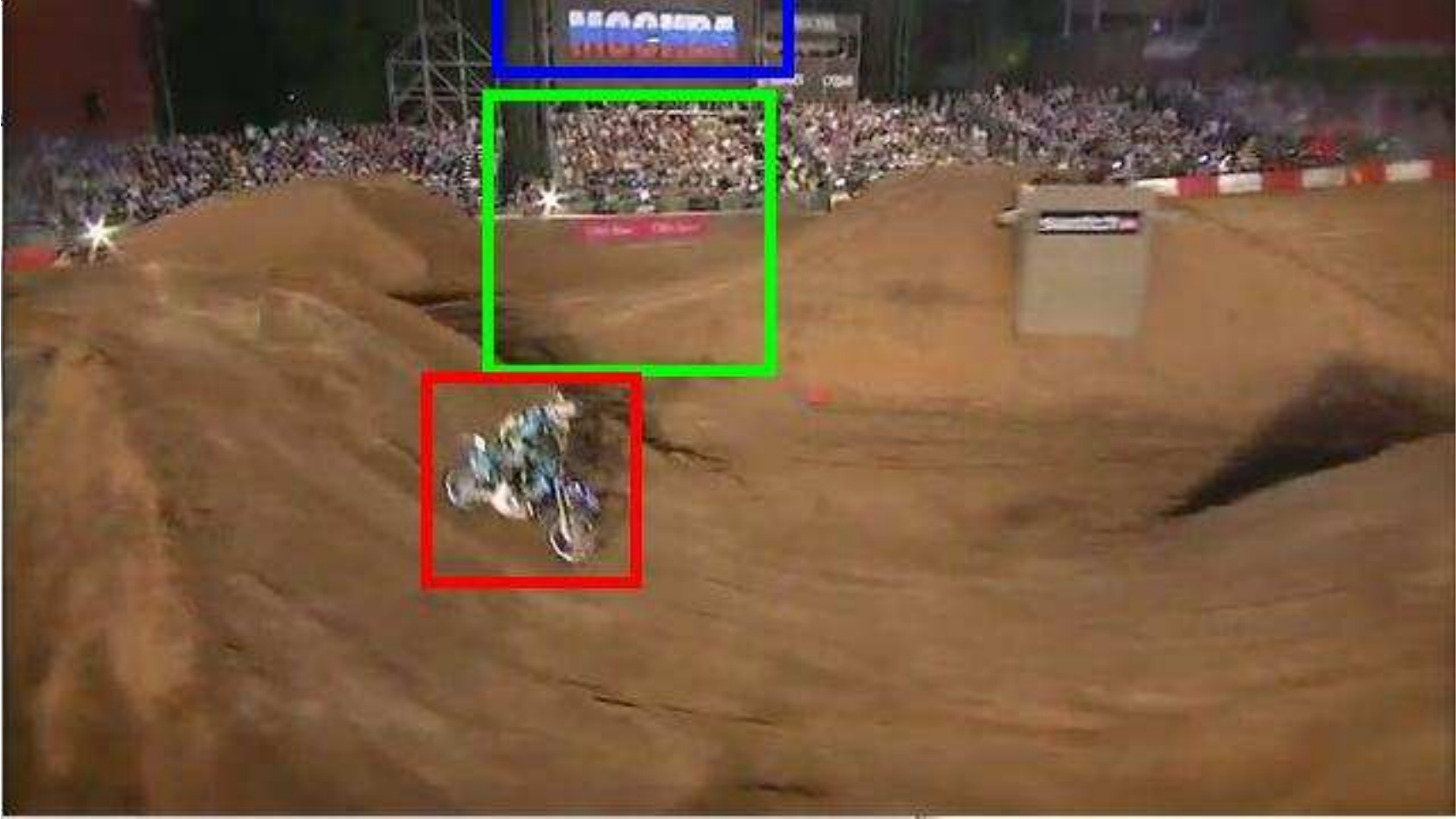}
	\includegraphics[width=0.32\linewidth]{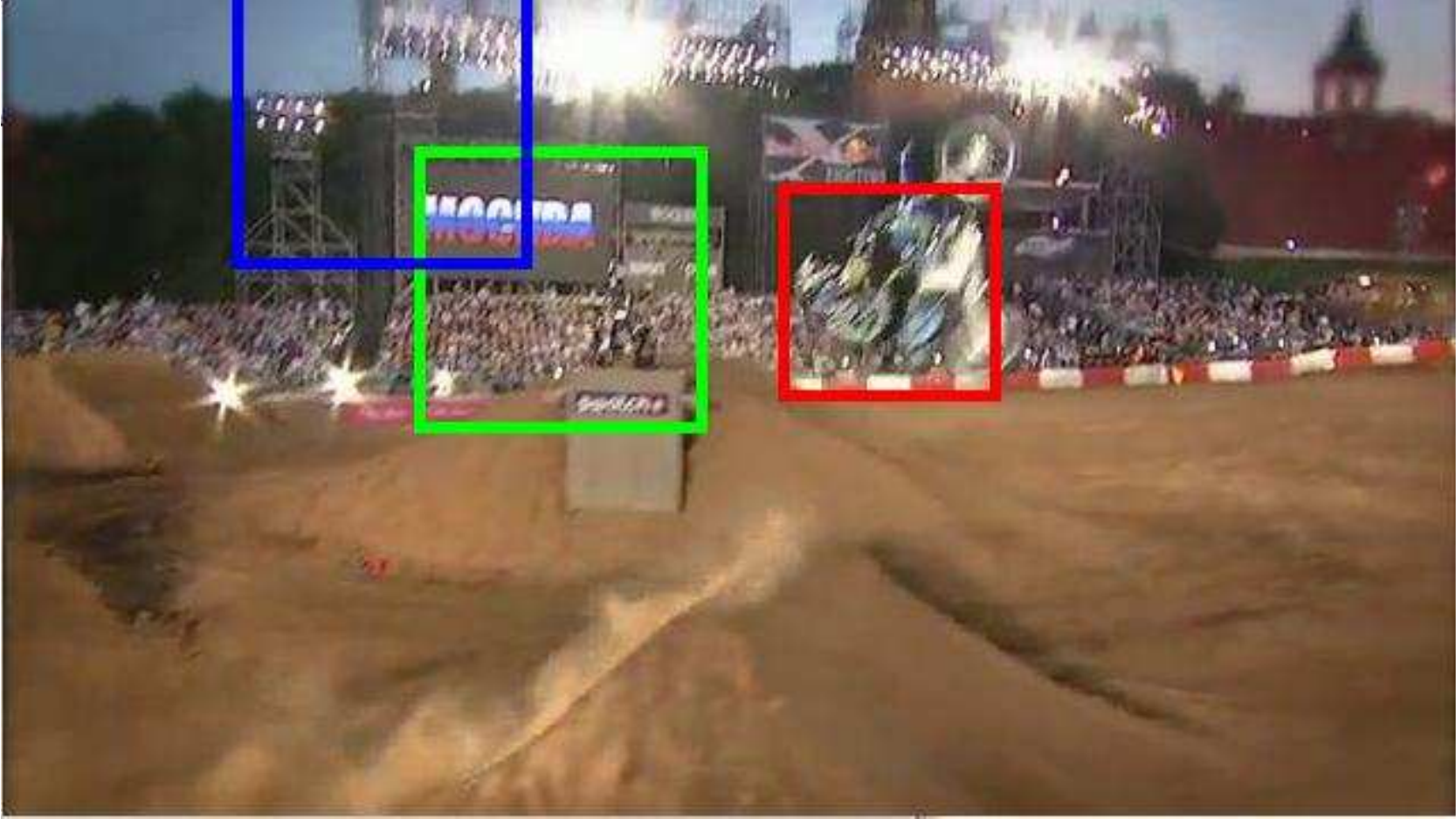}\\
	\includegraphics[width=0.32\linewidth]{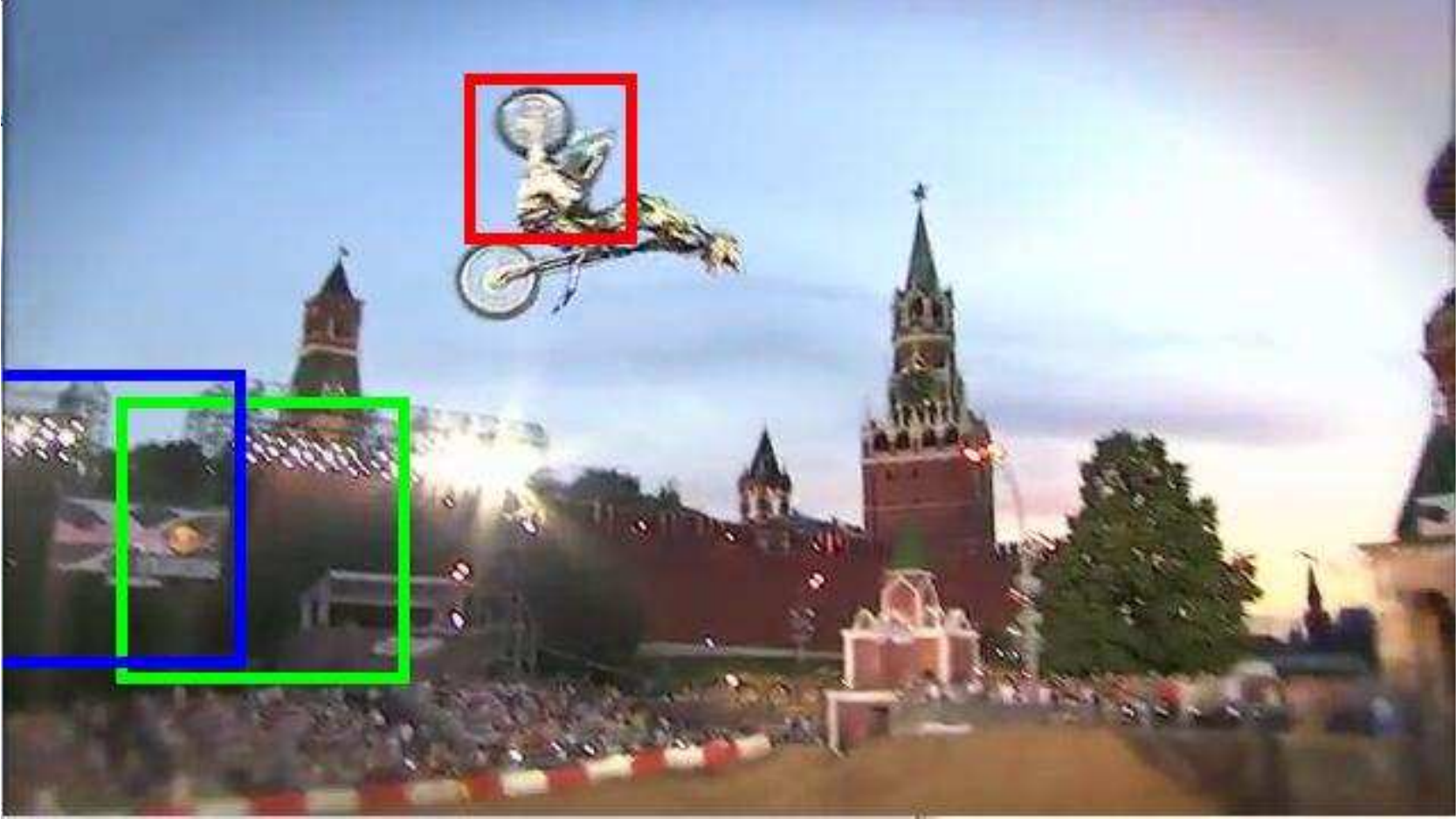}
	\includegraphics[width=0.32\linewidth]{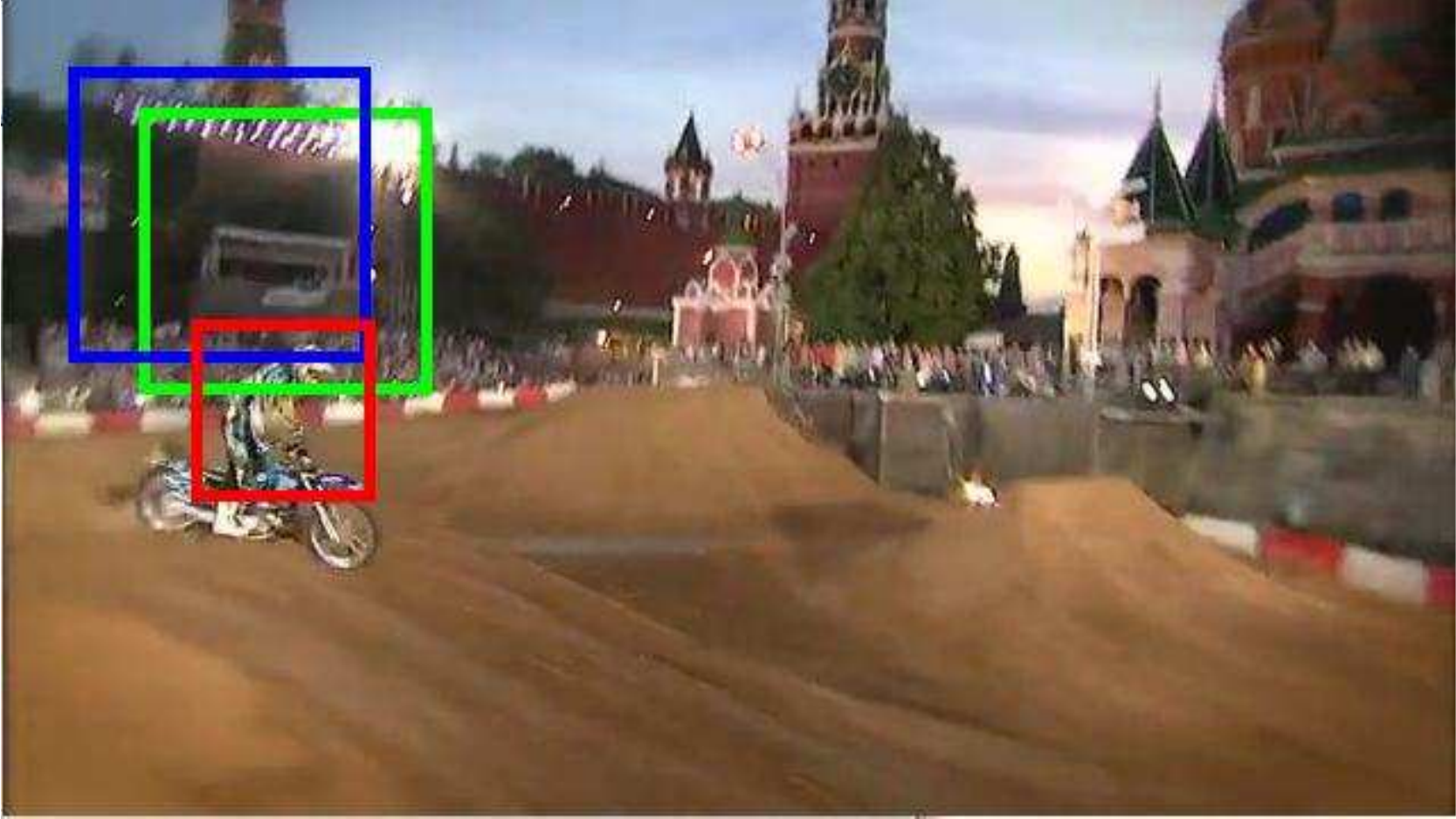}
	\includegraphics[width=0.32\linewidth]{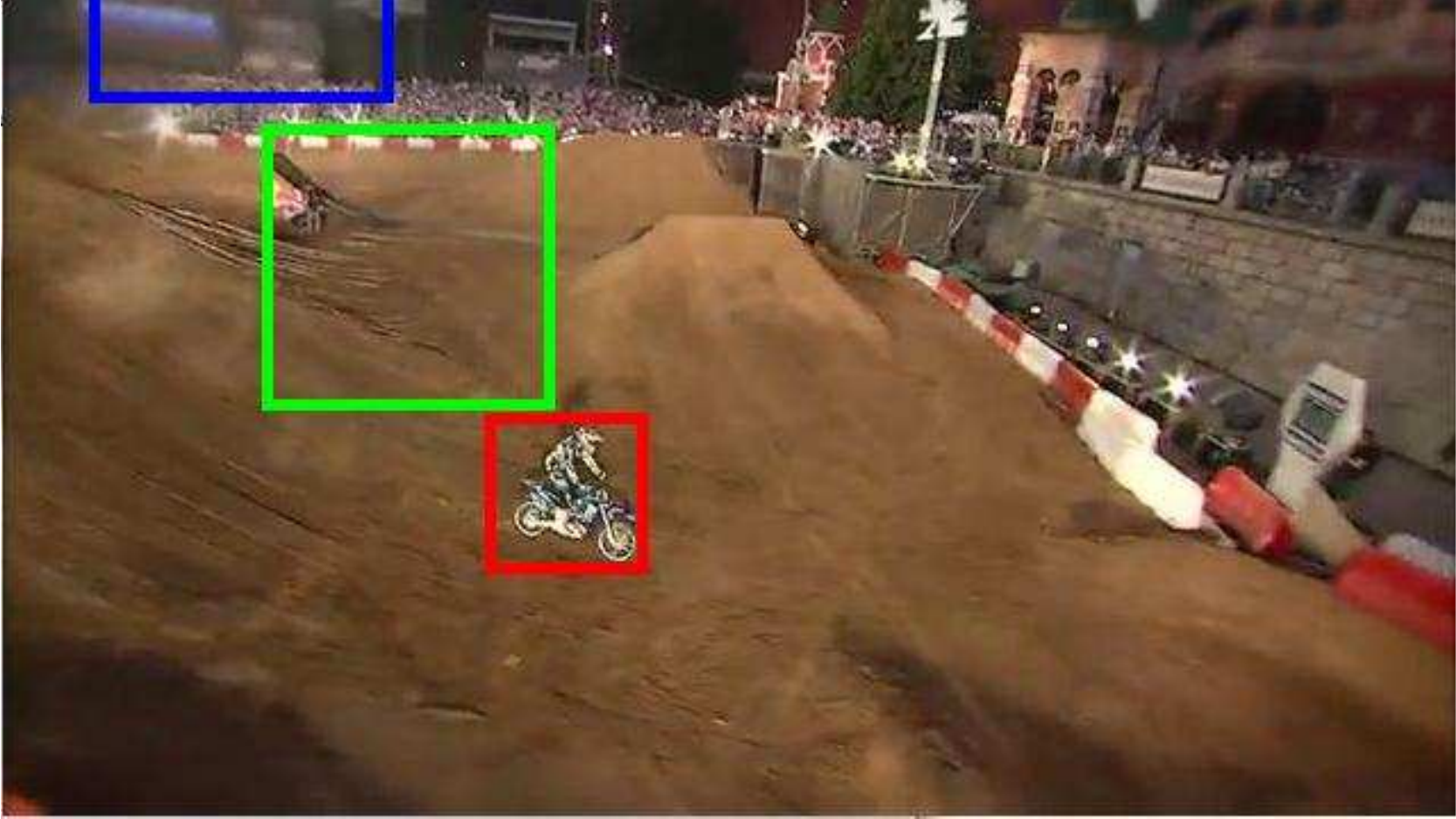}\\
	\vspace{2mm}
	\includegraphics[width=0.32\linewidth]{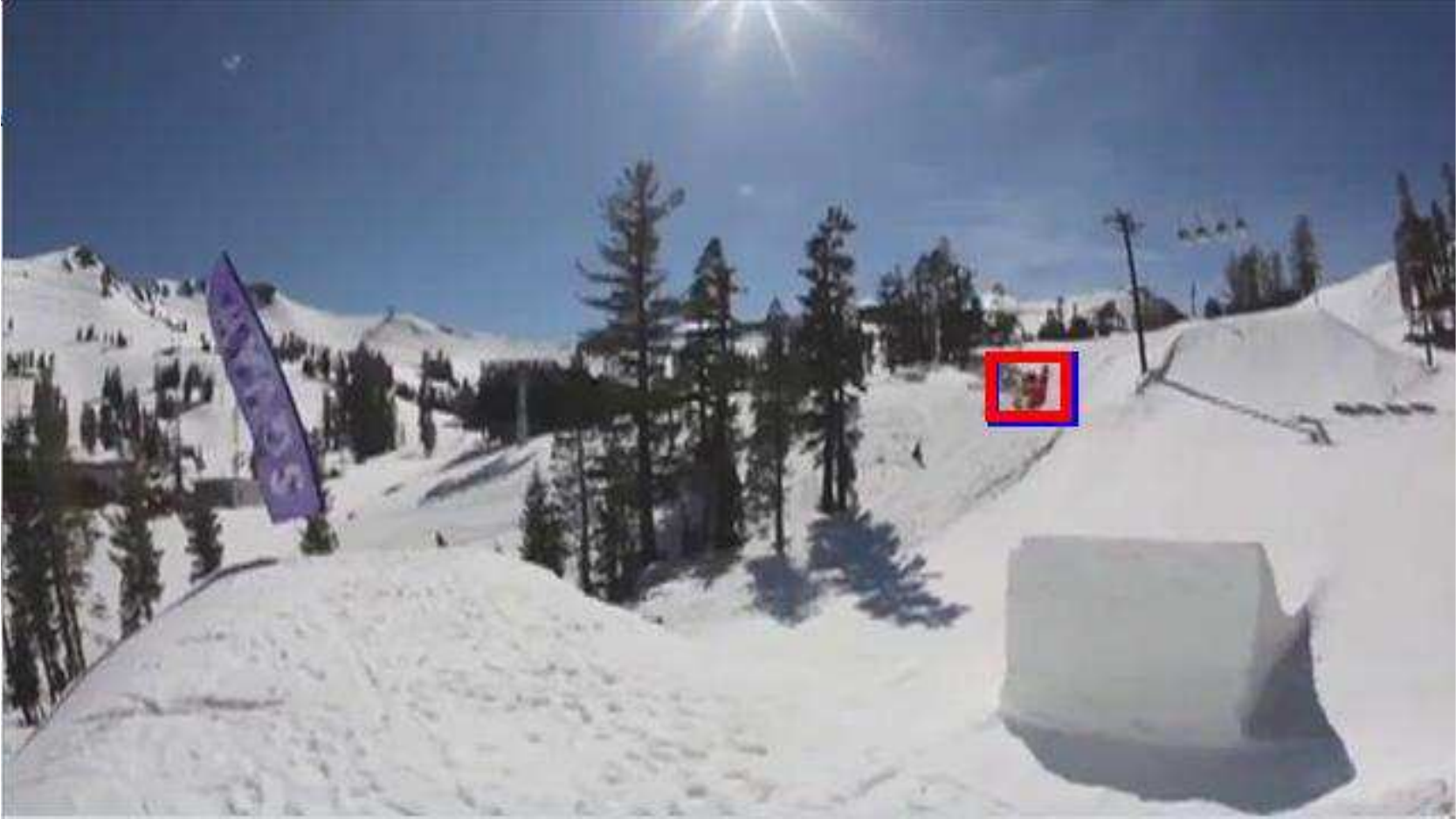}
	\includegraphics[width=0.32\linewidth]{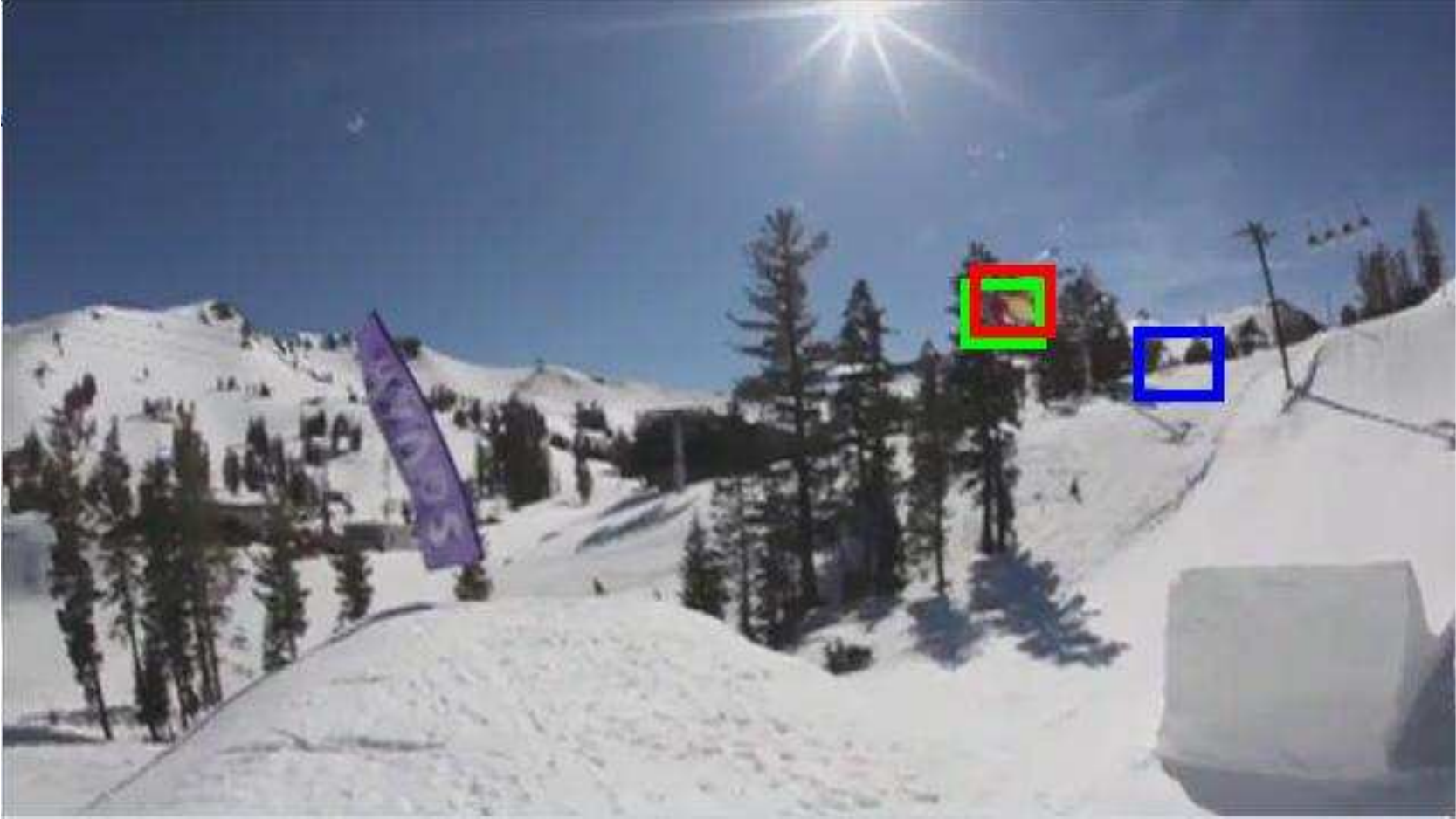}
	\includegraphics[width=0.32\linewidth]{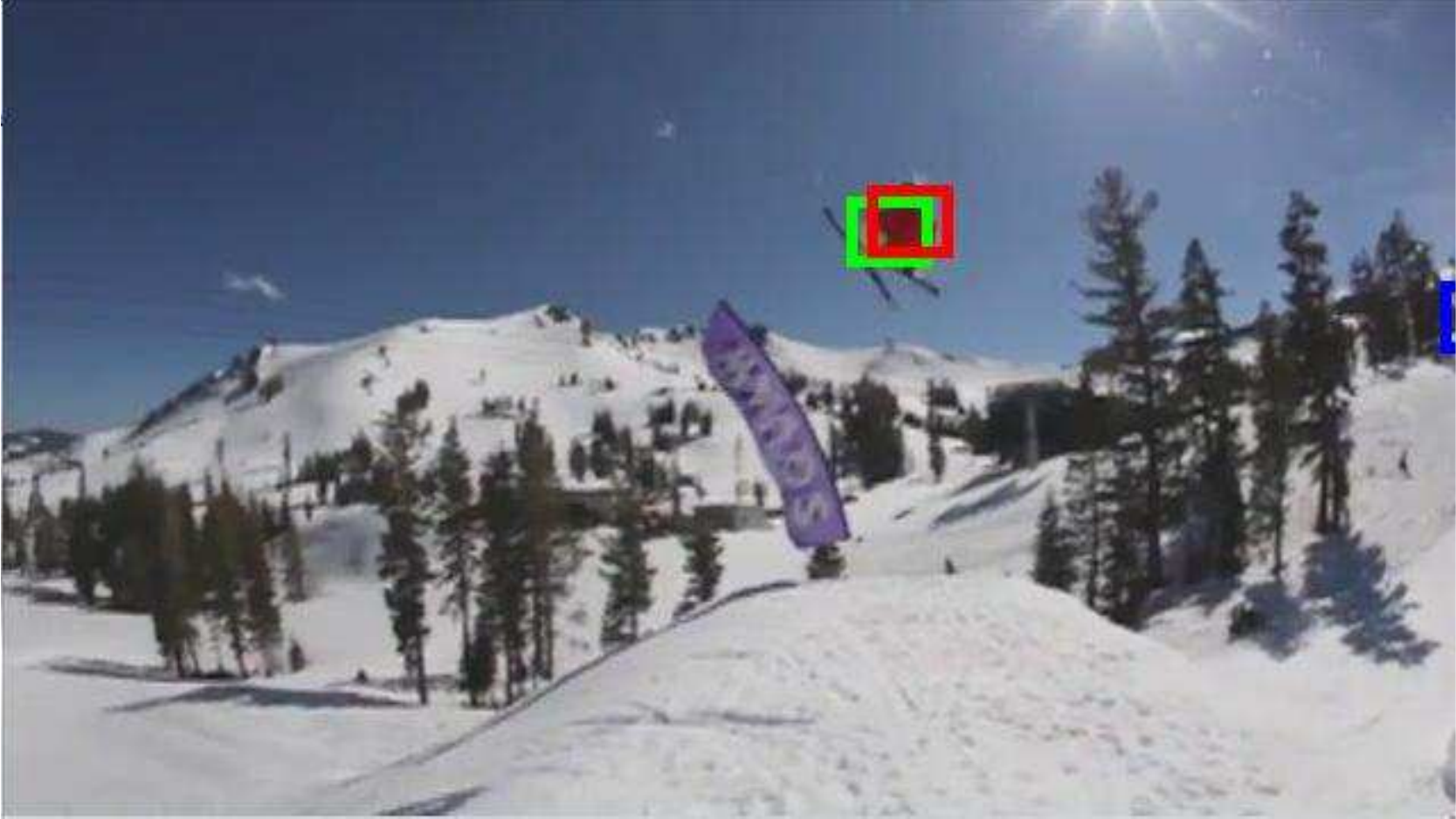}\\
	\includegraphics[width=0.32\linewidth]{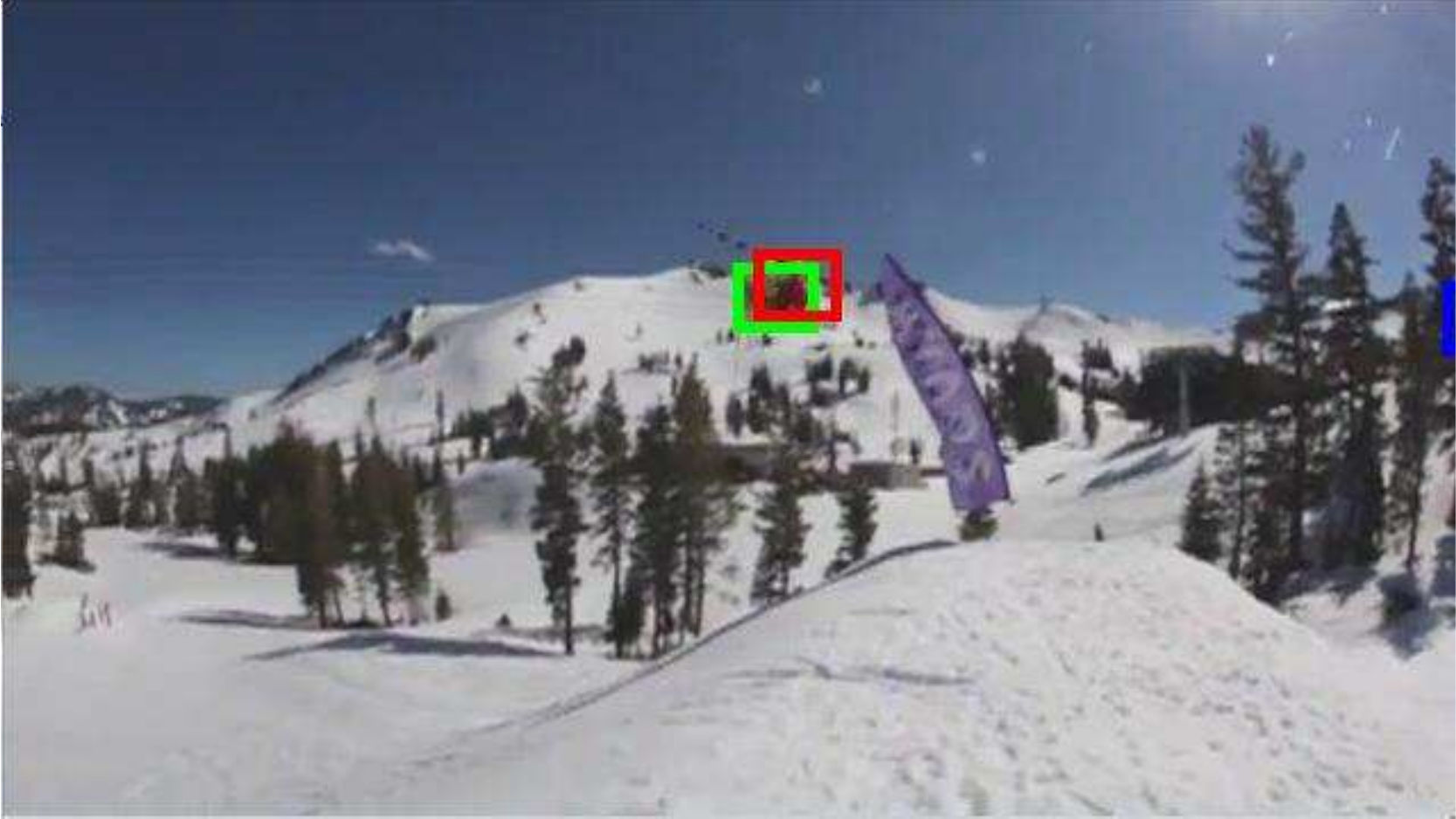}
	\includegraphics[width=0.32\linewidth]{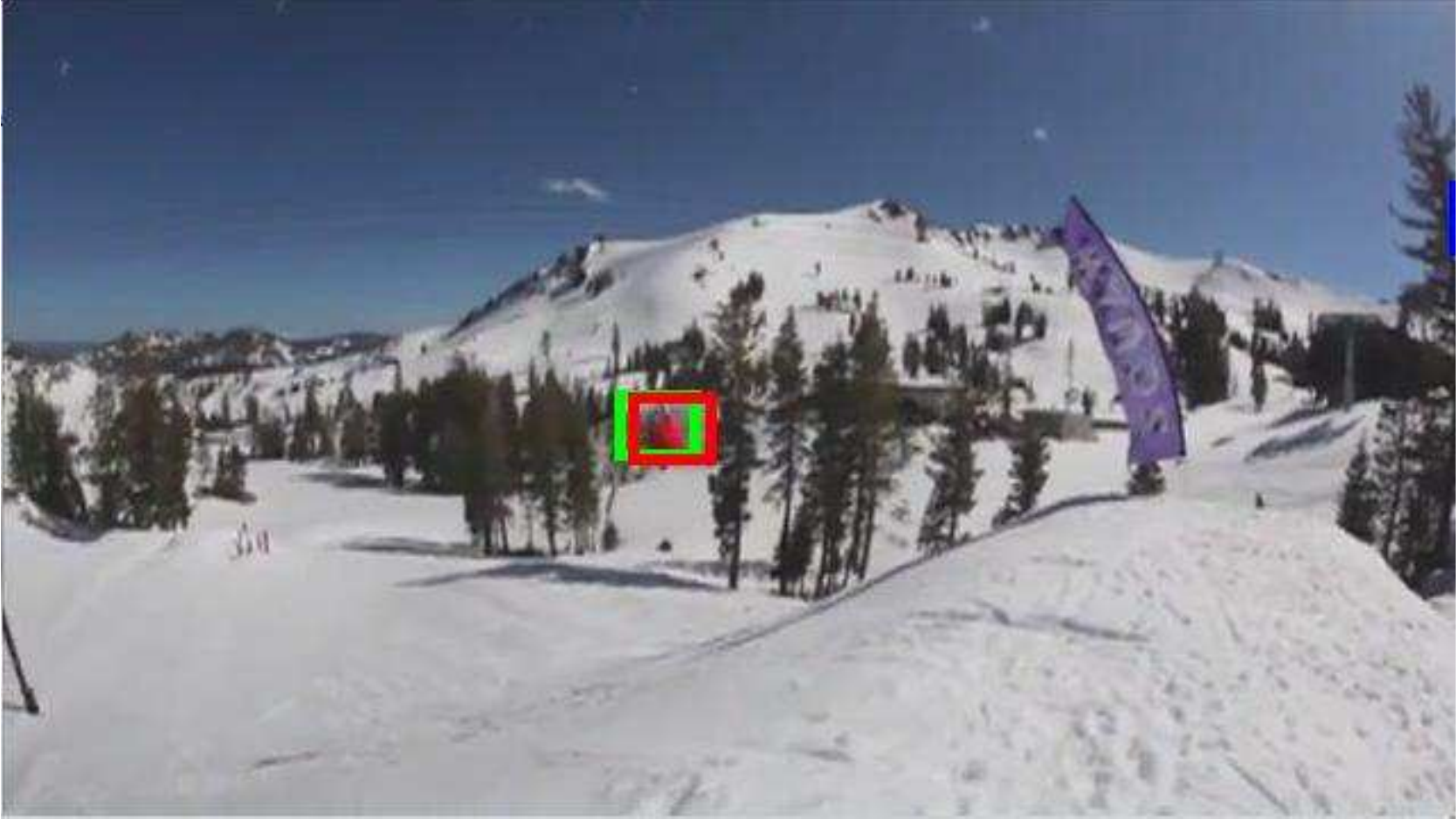}
	\includegraphics[width=0.32\linewidth]{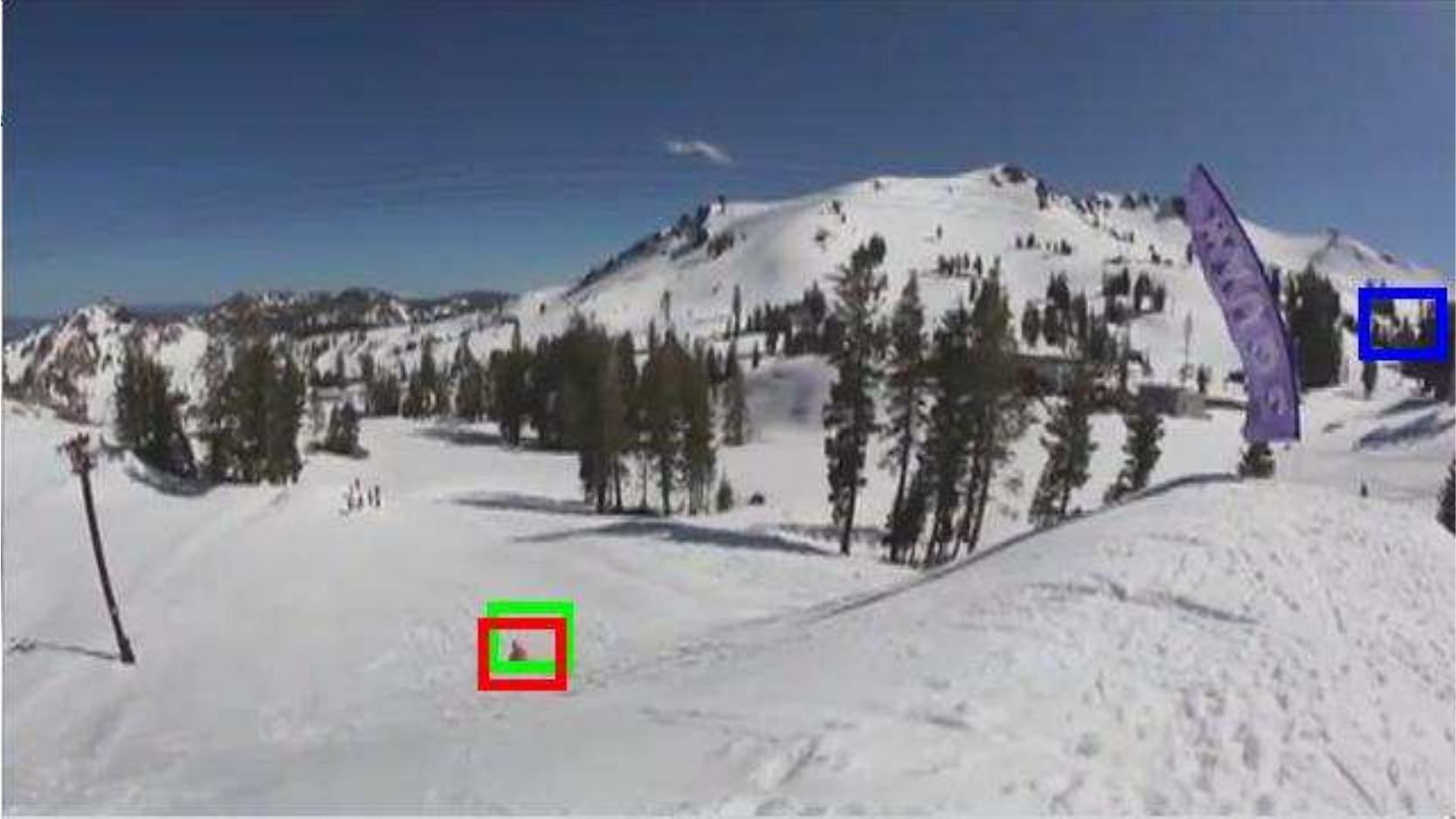}\\
	\vspace{2mm}
	\includegraphics[width=0.6\linewidth]{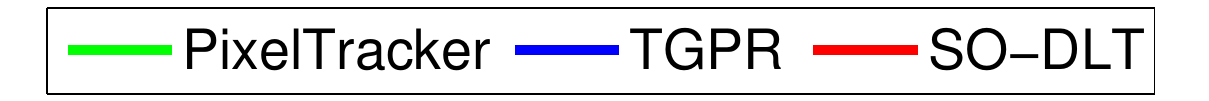}\\

	\caption{Tracking results for \emph{motocross1} and \emph{skiing} video sequences (SO-DLT is our proposed tracker).}\label{fig:teaser}
	}
\end{figure}

The past few years have been very exciting in the history of computer vision.  A great deal of excitement has been reported when applying the biologically-inspired convolutional neural network~(CNN) models to some challenging computer vision tasks.  For example, breakthrough performance has been reported for image classification~\cite{alexnet} and object detection~\cite{rcnn} tasks.  However, some other computer vision tasks such as visual tracking remain relatively unexplored in this recent surge of research interest.  We believe that a major reason is the lack of sufficient labeled training data which usually plays a very important role in achieving breakthrough performance for other applications because CNN training is typically done in a fully supervised manner.  In the case of visual tracking, however, labeled training data is usually very limited, often with only one labeled example as the object to track specified in the first frame of each video.  This makes direct application of the large-scale CNN approach infeasible.  In this paper, we present an approach which can address this challenge and hence can bring the CNN framework to visual tracking.  Using this approach to implement a tracker, we achieve very promising performance which outperforms the best state-of-the-art baseline tracker by more than 10\% (see Fig.~\ref{fig:teaser} for some qualitative tracking results).

Although visual tracking can be formulated in different settings according to different applications, the focus of this paper is the  \emph{one-pass model-free single object tracking} setting.  Specifically, it assumes that the bounding box of one single object in the first frame is given but no other appearance model is available.  Given this single (labeled) instance, the goal is to track the movement of the object in an online manner.  Consequently, this setting involves adapting the tracker to appearance changes of the object based on the possibly noisy output of the tracker.  Another way to formulate this problem would be as a self-taught one-shot learning problem in which the single example comes from the previous frame. Since learning a visual model from a single example is an ill-posed problem, a successful approach would require using some auxiliary data to learn an invariant representation of generic object features.  Although some recent work~\cite{dlt, wang2012transferring} also shares this spirit, the performance reported is inferior to the state of the art due to the lack of sufficient training data on one hand and the limited representational power of the model used on the other hand.  CNN has a role to play here by learning more robust features.  To make it feasible with limited training data during online tracking, we pre-train a CNN offline and then transfer the generic features learned to the online tracking task.

The first deep learning tracker (DLT)~\cite{dlt} reported in the literature is based on a stacked denoising autoencoder network.  While this approach is very promising, the exact realization of the approach reported in the paper has two limitations that hinder the tracking performance of DLT as compared to other state-of-the-art trackers.  First, the pre-training of DLT may not be very suitable for tracking applications.  The data used for pre-training is from the 80M Tiny Images dataset~\cite{tiny} with each image obtained by downsampling directly from a full-sized image.  Although some generic image features can be learned by learning to reconstruct the input images, the target to track in a typical tracking task is a single object rather than an entire image.  Features that are effective for tracking should be able to distinguish objects from non-objects (i.e.~background), not just to reconstruct an entire image.  Second, in each frame, DLT first generates candidates or proposals of the target based on the predictions of the previous frames, and then treats tracking as a classification problem. It ignores the structured nature of bounding boxes in that a bounding box or segmentation result corresponds to a region of an image, not just a simple label or real number as in a classification or regression problem.  Some previous work~\cite{struck, ebt} showed that exploiting the structured nature explicitly in the model could improve the performance significantly. Moreover, the number of proposals is usually in the order of several hundreds, making it hard to apply larger and more powerful deep learning models.

We propose a novel structured output CNN which transfers generic object features for online tracking. The contributions of our paper are summarized as follows:
\begin{enumerate}
	\item To alleviate the overfitting and drifting problems during online tracking, we pre-train the CNN to distinguish objects from non-objects instead of simply reconstructing the input or performing categorical classification on large-scale datasets with object-level annotations~\cite{imagenet}. 
	\item The output of the CNN is a pixel-wise map to indicate the probability that each pixel in the input image belongs to the bounding box of an object. The key advantages of the pixel-wise output are its induced structured loss and computational scalability.
	\item We evaluate our proposed method on an open benchmark~\cite{benchmark} as well as a challenging non-rigid object tracking dataset and obtain very remarkable results.  In particular, we improve the \emph{area under curve} (AUC) metric of the overlap rate curve from 0.529 to 0.602 for the open benchmark.
\end{enumerate}



\section {Related Work}
\subsection{Deep Learning and CNNs}
The root of deep learning can be dated back to research on multilayered neural networks in the late 1980s. The resurgence of research interest in neural networks owes to a more recent work~\cite{hintonscience} which used pre-training to make the training of deeper networks feasible.  Among different deep learning models, CNN seems to be a more suitable choice for many vision tasks as the design of CNN has been inspired by the vision systems of the biological counterparts.  Among its characteristics, the convolution operation can capture local and repetitive similarity and the pooling operation can allow local translational invariance in images.  The rapid development of powerful computing devices such as \emph{general-purpose graphics processing units} (GPGPU) and the availability of large-scale labeled datasets such as ImageNet~\cite{imagenet} have made the training of large-scale CNN possible.  It has been demonstrated visually in~\cite{zeiler2014visualizing} that a CNN can gradually learn low-level to high-level features through the transformation and enlargement of receptive fields in different layers.  As opposed to using handcrafted features as in the conventional recognition pipeline, it has been demonstrated that the features learned by a large-scale CNN can achieve very superior performance in some high-level vision tasks such as image classification~\cite{alexnet} and object detection~\cite{rcnn}.

\subsection{Visual Tracking}
Many methods have been proposed for single object tracking. For a systematic review and comparison, we refer the readers to a recent survey and a benchmark~\cite{survey, benchmark}. Most of the existing tracking methods belong to the general framework of Bayesian tracking~\cite{tutorial}. It decomposes the problem into two parts which involve a motion model and an appearance model.  Although some trackers attempt to go beyond this framework, e.g.~\cite{vtd, mug, interval}, most still focus on improving the appearance model because this aspect is crucial to enhancement in performance.

Generally speaking, most trackers belong to either one of two categories: generative trackers and discriminative trackers. Generative trackers usually assume a generative process of the tracked target and search for the most probable candidate as the tracking result. Some representative methods are based on principal component analysis~\cite{ivt}, sparse coding~\cite{l1t}, and dictionary learning~\cite{onndl}. On the other hand, discriminative trackers learn to separate the foreground from the background using a classifier. Many advanced machine learning algorithms have been used, including boosting variants~\cite{oab, semiB}, multiple-instance learning~\cite{mil}, structured output SVM~\cite{struck}, and Gaussian process regression~\cite{gpr}.  These two approaches are in general complementary.  Discriminative trackers are usually more resistant to cluttered background since they explicitly sample image patches from the background as negative training examples.  On the other hand, generative trackers are usually more accurate under normal situations.  Besides, some methods exploit correlation filters for the target or context~\cite{mosse, kcf, stc}. Their primary advantage is that only fast Fourier transform and several matrix operations are needed, making them very suitable for real-time applications.  Moreover, some methods take the ensemble learning approach~\cite{bailer2014superior, meem, ebt} which is especially effective when the constituent trackers involved in the ensemble have high diversity.  Furthermore, some methods focus on long-term tracking, e.g.~\cite{tld, selfpaced}.

As for applying deep learning to visual tracking, besides the DLT~\cite{dlt} mentioned in the previous section, some recent methods include using an ensemble~\cite{dltensemble} and maintaining a pool of CNNs~\cite{deeptrack}. However, due to the lack of sufficient training data, these methods only show comparable or even inferior results compared to other state-of-the-art trackers. In summary, for visual tracking applications, we believe that the power of deep learning has not yet been fully liberated.

\section{Our Tracker}
In this section, we will present our \emph{structured output deep learning tracker} (SO-DLT). We first present the CNN architecture in SO-DLT and the offline pre-training process of the CNN.  We then present details of the online tracking process.

\subsection{Overview}

Training of the tracker can be divided into two stages, the offline pre-training stage and the online fine-tuning and tracking stage.  In the pre-training stage, we train a CNN to learn generic object features for distinguishing objects from non-objects, i.e., to learn from examples the notion of \emph{objectness}.  Instead of fixing the learned parameters of CNN during online tracking, we fine-tune them so that the CNN can adapt to the target being tracked.  For robustness, we run two CNNs concurrently during online tracking to account for possible mistakes caused by model update.  The two CNNs work collaboratively in determining the tracking result of each video frame.

\subsection{Objectness Pre-training}
\begin{figure*}[htb]
	\center{
	\includegraphics[width=0.95\linewidth]{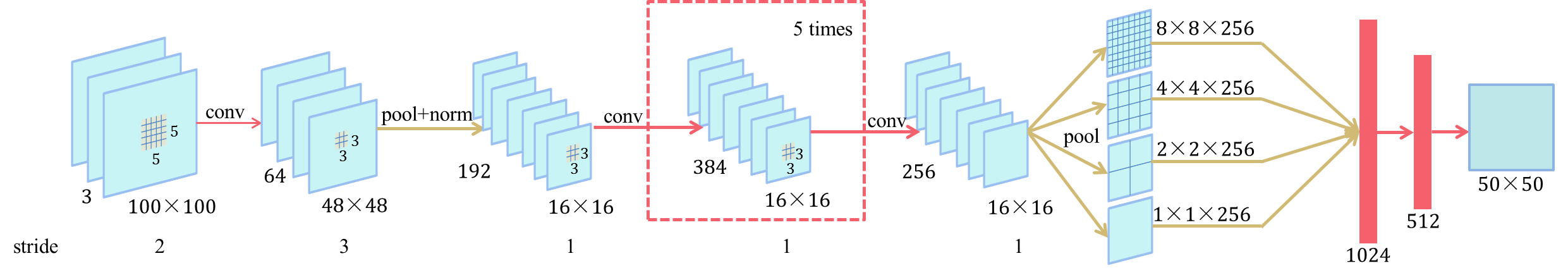}
	\caption{Architecture of the proposed structured output CNN.}\label{fig:cnn}
	}
\end{figure*}

The architecture of the structured output CNN is shown in Fig.~\ref{fig:cnn}. It consists of seven convolutional layers and three fully connected layers. Between these two parts, a multiscale pooling scheme~\cite{spp} is introduced to retain more features related to locality since the output needs them for localization.
The parameter setting of the network is shown in Fig.~\ref{fig:cnn}. In contrast to the conventional CNN used for classification or regression, there is a crucial difference in our model: the output of the CNN is a $50 \times 50$ probability map rather than a single number.
Each output pixel corresponds to a $2 \times 2$ region in the original input, with its value representing the probability that the corresponding input region belongs to an object. In our implementation, the output layer is a 2500-dimensional fully connected layer which is then reshaped to the $50 \times 50$ probability map. Since there exists strong correlation between neighboring pixels of the probability map, we only use 512 hidden units in the previous layer to help prevent overfitting.

To train such a large CNN, it is essential to use a large dataset to prevent overfitting.  Since we are interested in object-level features, we use the ImageNet 2014 detection dataset\footnote{\url{http://image-net.org/challenges/LSVRC/2014/}} which contains 478,807 bounding boxes in the training set. For each annotated bounding box, we add random padding and scaling around it. We also randomly sample some negative examples when the overlap rates\footnote{The overlap rate between two bounding boxes is defined as the area of intersection of the two bounding boxes over the area of their union.} of the positive examples are below a certain threshold. Note that it does not learn to distinguish different object classes as in a typical classification or detection task, since we are only interested in learning to differentiate objects from non-objects in this stage.  Consequently, we use an element-wise logistic regression model in each position of the $50 \times 50$ output map and define the loss function accordingly.  For the training target, a pixel inside the bounding box is set to 1 while it is 0 outside.  As for a negative example, the target is 0 for the entire probability map. This setting is equivalent to penalizing the number of mismatched pixels between the prediction and the ground truth, thus inducing a structured loss function which fits the problem better. Mathematically, let $p_{ij}$ denotes the prediction of $(i, j)$ position, and $t_{ij}$ is a binary variable denotes the ground truth of $(i, j)$ position, the loss function of our method is defined as:
\begin{equation}
	\min_{p_{ij}} \sum_{i = 1}^{50} \sum_{j = 1}^{50} -(1 - t_{ij}) \log (1 - p_{ij}) - t_{ij} \log(p_{ij}).
\end{equation}
The detailed parameters for training are described in Sec.~\ref{sec:impl}.

\begin{figure}[htb]
	\center{
	\includegraphics[width=0.85\linewidth]{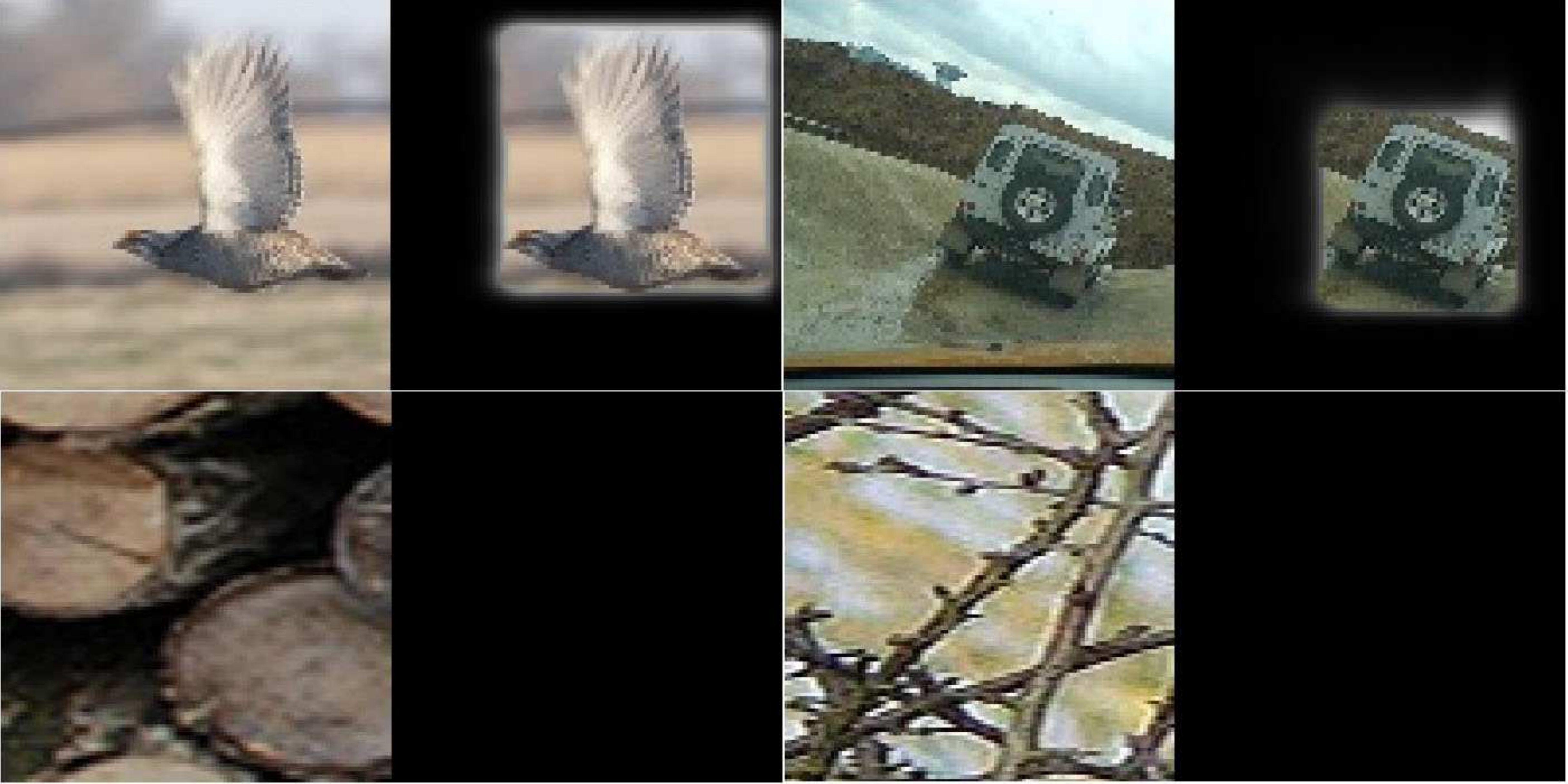}
	\vspace{2mm}
	\caption{Testing of the pre-trained objectness CNN on the ImageNet 2014 detection validation set. The first row shows two positive examples each of which contains an object. The objectness CNN can accurately detect the object position and scale. The second row shows two negative examples. The objectness CNN does not fire on them showing the lack of evidence for the existence of any object of interest. The CNN plays an important role in making our SO-DLT robust to occlusion and cluttered background during online tracking.}\label{fig:pretrain}
	}
\end{figure}

Fig.~\ref{fig:pretrain} shows some results when the pre-trained CNN is tested on the held-out validation set provided by the ImageNet 2014 detection task. In most cases, the CNN can successfully determine whether the input image contains an object, and if yes it can accurately locate the object of interest. Note that since the labels of our training data are only bounding boxes, the output of the $50 \times 50$ probability map is also in the form of a square.  Although there are methods~\cite{boxsup} that utilize bounding box information to provide weak supervision and obtain pixel-wise segmentation as well, we believe that the probability map output in our model is sufficient for the purpose of tracking.

\subsection{Online Tracking}

The CNN pre-trained to learn generic object features as described above cannot be used directly for online tracking because the data bias of the ImageNet data is different from that of the data observed during online tracking.  Moreover, if we do not fine-tune the CNN, it will fire on all objects that appear in a video frame instead of just the object being tracked.  Therefore, it is essential to fine-tune the pre-trained CNN using the annotation in the first frame of each video collected during online tracking to make sure that the CNN is specific to the target.  Note that fine-tuning, or online model adaptation, is an indispensable part of our tracker rather than an optional feature solely introduced to further improve the tracking performance.

\begin{figure*}[htb]
	\center{
	\includegraphics[width=0.88\linewidth]{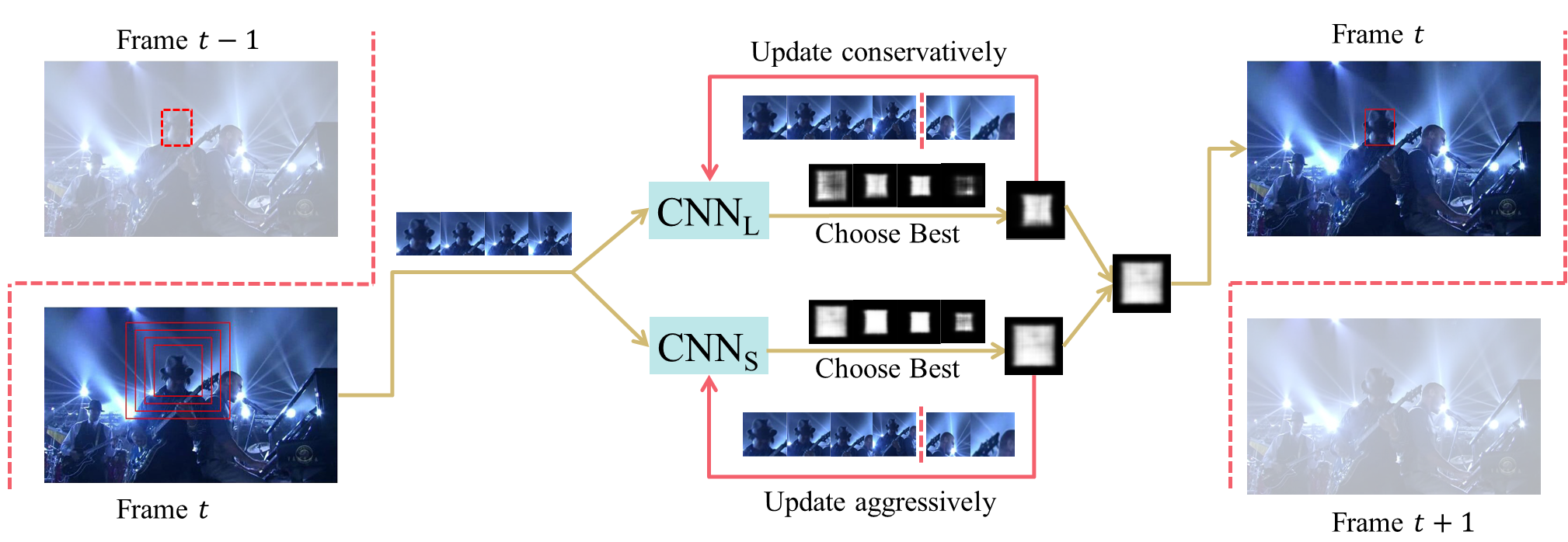}
	\caption{Pipeline of our tracking algorithm.}\label{fig:pipeline}
	}
\end{figure*}

We now present the basic online tracking pipeline.  We maintain two CNNs which use different model update strategies. After fine-tuning using the annotation in the first frame, we crop some image patches from each new frame based on the estimation of the previous frame. By making a simple forward pass through the CNN, we can obtain the probability map for each of the image patches. The final estimation is then determined by searching for a proper bounding box. The two CNNs are updated if necessary. We illustrate the pipeline of the tracking algorithm in Fig.~\ref{fig:pipeline}. In what follows, we will elaborate the major steps of the pipeline separately.

\subsubsection{Bounding Box Determination}

When a new frame comes, the first step of our tracker is to determine the best location and scale of the target.
We first specify the possible regions that may contain the target and feed the regions into the CNN.  Next, we decide the most probable location of the bounding box based on the probability map.

\noindent
\textbf{Search Mechanism:} Selecting a proper search range for the target is a nontrivial problem. Using too small search regions makes it easy to lose track of a target under fast motion, but using too large search regions may include salient distractors in the background. For example, in Fig.~\ref{fig:sample}, the output response gets weaker as the search region is enlarged mainly due to the cluttered background and another person nearby. To address this issue, we propose a multi-scale search scheme for determining the proper bounding box. First, all the cropped regions are centered at the estimation of the previous frame. Then, we start searching with the smallest scale. If the sum over the output probability map is below a threshold (i.e., the target may not be in this scale), then we proceed to the next larger scale.  If we cannot find the object in all scales, we report that the target is missing.

\noindent
\textbf{Generating Bounding Box:}
After we have selected the best scale, we need to generate the final bounding box for the current frame. We first determine the center of the bounding box and then estimate its scale change with respect to the previous frame. To determine the center we use a density based method, which sets a threshold $\tau_1$ for the corresponding probability map and finds a bounding box with all probability values inside above the threshold.  Next, the bounding box location under the current scale is estimated by taking an average over the different values of $\tau_1$.
After the center is determined, we need to search again in the corresponding region to find a proper scale. The scale is aimed at fitting the exact target region perfectly. Simply using the average confidence (which makes the tracker prefer the central area with high confidence) or total confidence (which makes it prefer the whole frame) is not satisfactory.

Let $P$ denote the output probability map and $p_{ij}$ the $(i, j)$th element in $P$.  We consider a bounding box with top-left corner $(x,y)$, width $w$ and height $h$.  Its score is calculated as
\begin{equation}
	c = \sum_{i = x}^{x + w - 1} \sum_{j = y}^{y + h - 1} (p_{ij} - \epsilon) \cdot w \cdot h,
\end{equation}
where $\epsilon$ balances the scale of the bounding box. We also repeat with several $\epsilon$ values and average their results for robust estimation. The confidence can be calculated very efficiently with the help of integral images~\cite{integral}.

\subsubsection{Differentially-paced Fine-tuning}

Model update in visual tracking often faces a dilemma.  If the tracker does not update frequently enough, it may not adapt well to appearance changes.  However, if it updates too frequently, inaccurate results may impair its performance and lead to the drifting problem.

We tackle this dilemma by using two CNNs during online tracking. The basic idea is to make one CNN ($\text{CNN}_S$) account for short-term appearance while the other one ($\text{CNN}_L$) for long-term appearance. First, both CNNs are fine-tuned in the first frame of a video. Afterwards, $\text{CNN}_L$ is tuned conservatively while $\text{CNN}_S$ is tuned aggressively. By working collaboratively,  $\text{CNN}_S$ adapts to dramatic appearance changes while $\text{CNN}_L$ is resistant to potential mistakes. The final estimation is then determined by the more confident one. Consequently, the final integrated results are more robust to drifting caused by occlusion or a cluttered background. Although there exist more advanced model update methods, e.g.~\cite{meem}, we find that this simple scheme works quite well in practice. 

We now provide more details about the update strategies. We first observe that if a model is updated immediately as soon as the prediction falls below a threshold, the model will be easily influenced by noisy results. On the other hand, we find that the quality of negative examples is generally quite stable. 
As a result, $\text{CNN}_S$ is updated when there exists a negative example such that
\begin{equation}
	\sum_{i = 1}^{50} \sum_{j = 1}^{50} p_{ij} > \tau_2.
\end{equation}
This is to make sure that any background object that makes the CNN fire should be suppressed.  Doing so will reduce the chance that the tracker will drift towards some negative examples similar to the tracked object when the following frames are processed.
On the contrary, in addition to the above condition, $\text{CNN}_L$ will only be updated if
\begin{equation}
	\sum_{i = x}^{x + w - 1} \sum_{j = y}^{y + h - 1} p_{ij} > \tau_3 \cdot w \cdot h, 
\end{equation}
where $(x, y, w, h)$ denotes the output target bounding box in the current frame. This means that we update $\text{CNN}_L$ more conservatively in that we only update it if we are highly confident about the results in the current frame. Doing so will reduce the risk of incorrect update when the true target has already drifted to the background.

\begin{figure}[htb]
	\center{
	\subfigure[]{\includegraphics[width=0.6\linewidth]{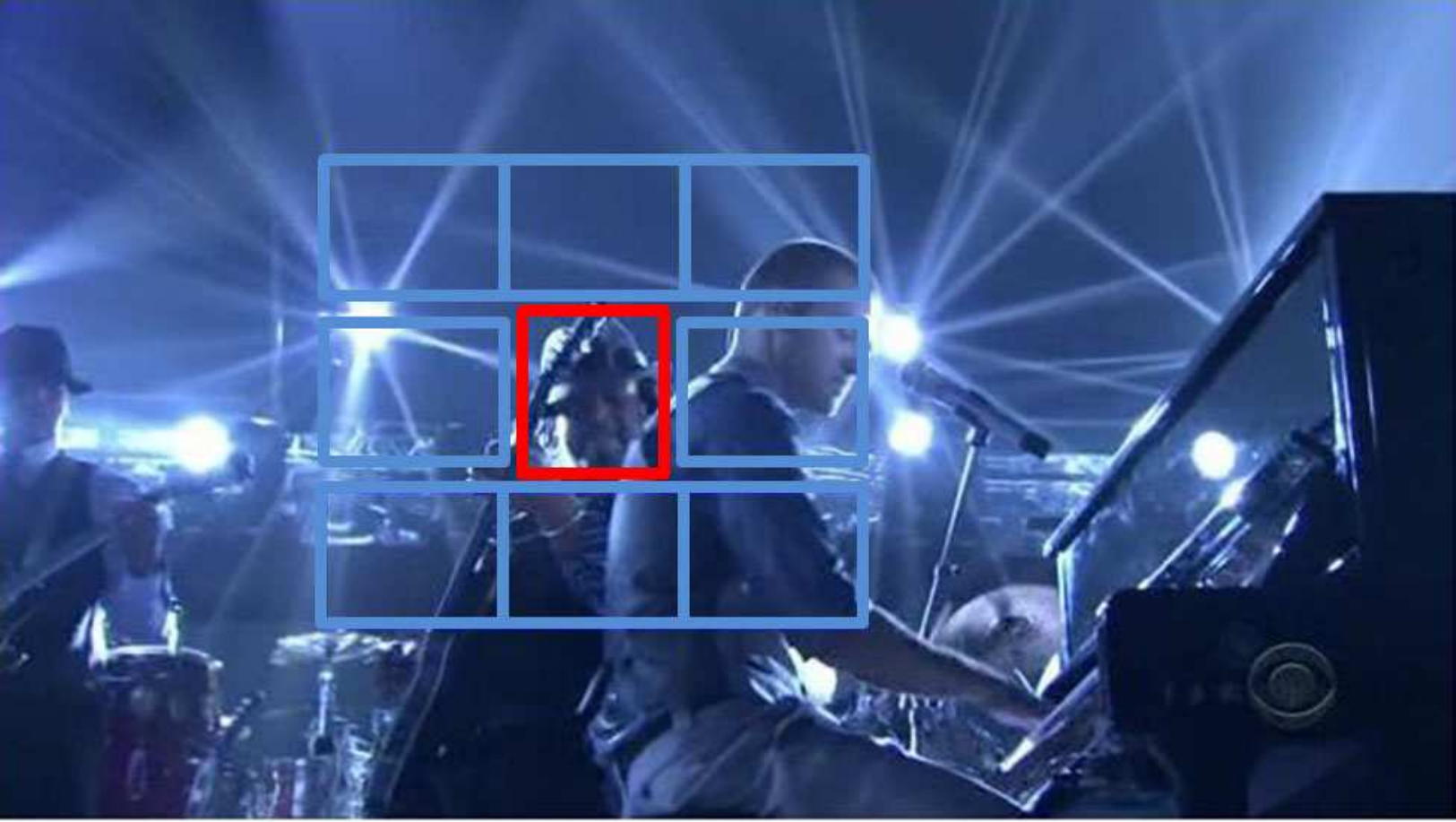}}~~~
	\subfigure[]{\includegraphics[width=0.32\linewidth]{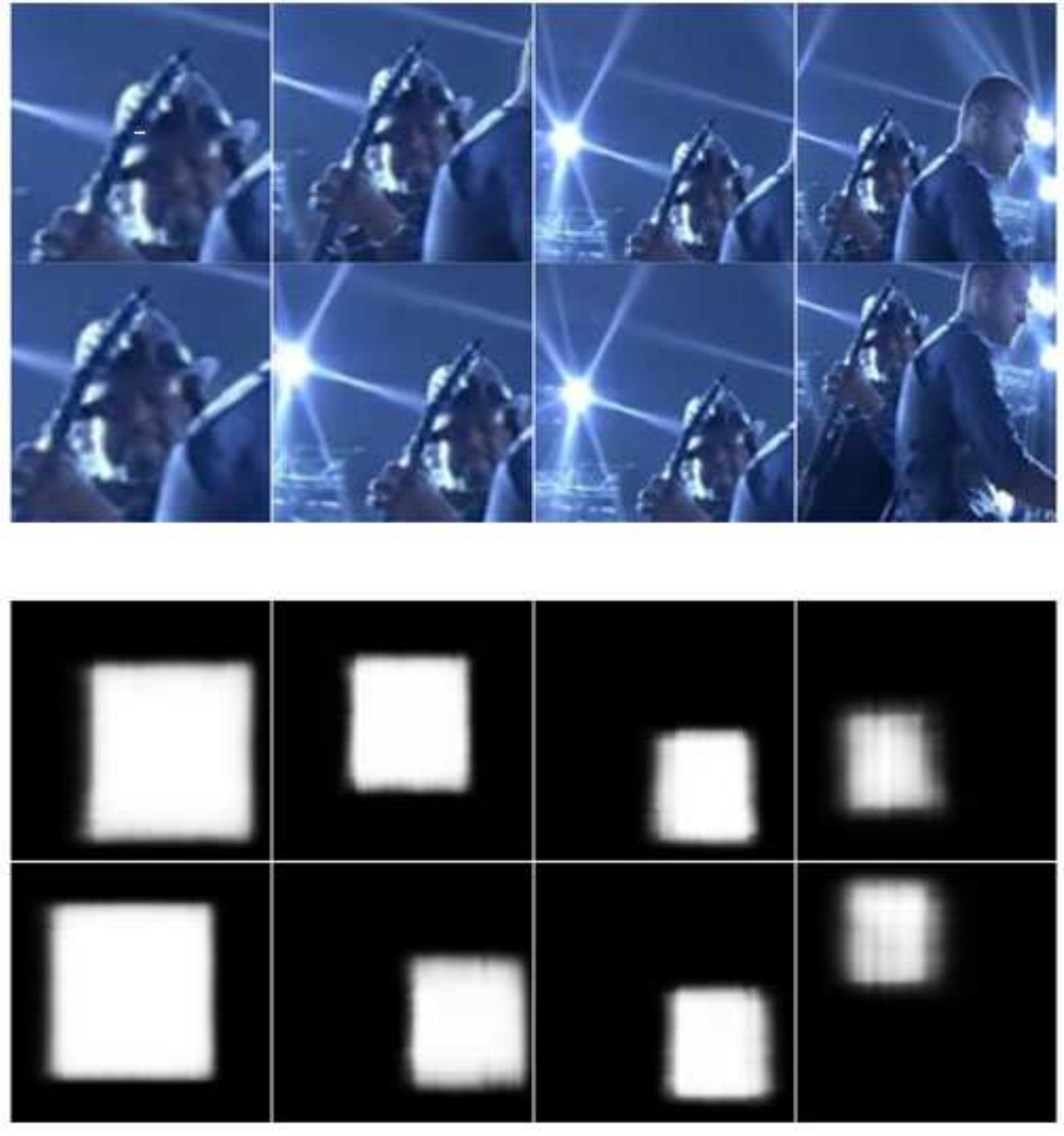}}
	\caption{Sampling scheme of the proposed tracker. On the left, the red bounding box denotes the target to track while the eight blue ones around it are negative examples. On the right, we show in the upper part positive examples fed into the CNN. They are padded with different scales and random translations.  The lower part shows the corresponding output of the CNN after applying fine-tuning to this frame.}\label{fig:sample}
	}
\end{figure}

In each update, we need to collect both positive and negative examples. Our sampling scheme is illustrated in Fig.~\ref{fig:sample}. For positive examples, we sample them in four scales based on the estimation of the previous frame. Random translation is also introduced to eliminate the learning bias to the center location. As for negative examples, we crop eight non-overlapping bounding boxes around the target in different directions in two scales. The output of the positive examples is also shown in Fig~\ref{fig:sample}.

\section{Experiments}
In this section, we empirically validate the proposed SO-DLT tracker by comparing it with other state-of-the-art trackers.  For fair comparison, not only do we need a reasonably large benchmark dataset to avoid bias due to data selection, but there should also be a carefully designed protocol that is followed by every tracker.  A recent work~\cite{benchmark} introduced a unified tracking benchmark which includes both a dataset and a protocol.  We use the benchmark dataset for our comparative study and strictly follow the protocol by fixing the same set of parameters for all video sequences tested.  We will make our implementation publicly available if the paper is accepted.

\subsection{Implementation Details}
\label{sec:impl}
The part related to CNN is implemented using the Caffe toolbox~\footnote{\url{http://caffe.berkeleyvision.org}} and the wrapper for online tracking is implemented directly in MATLAB. All the experiments are run on a desktop computer with a 3.40GHz CPU and a K40 GPU. The speed of our unoptimized code is about 4~to~5 frames per second.


For the pre-training of CNN, we start with a learning rate of $10^{-7}$ with momentum $0.9$ and decrease the learning rate once every 5 epochs. We train for about 15 epochs in total. Note that our learning rate is much smaller than typical choices due to the different loss function used by us. To alleviate overfitting, a weight decay of  $5 \times 10^{-4}$ is used for each layer and the first fully connected layer is regularized with a dropout rate of $0.5$. During fine-tuning, we use a larger learning rate of  $2 \times 10^{-7}$with a smaller momentum of $0.5$. For the first frame, we fine-tune each CNN for 20 iterations. For subsequent frames, we only fine-tune for one iteration.

 $\tau_1$ ranges from $0.1$ to $0.7$ with a step size $0.05$. The threshold of sum of confidence $\tau_2$ for negative examples is set to $\tau_2 = 100 $. The updating threshold of $\text{CNN}_L$ is set to $\tau_3 = 0.8$. The normalized constant $\epsilon$ for searching proper scales ranges from $0.55$ to $0.6$ with a step size of $0.025$.

\subsection{CVPR2013 Visual Tracker Benchmark}
The CVPR2013 Visual Tracker Benchmark~\cite{benchmark} contains 50 fully annotated sequences and covers a  variety of challenging scenarios in the tracking literature over the past few years. 
\begin{figure}[htb]
	\centering
	\includegraphics[width=0.47\linewidth]{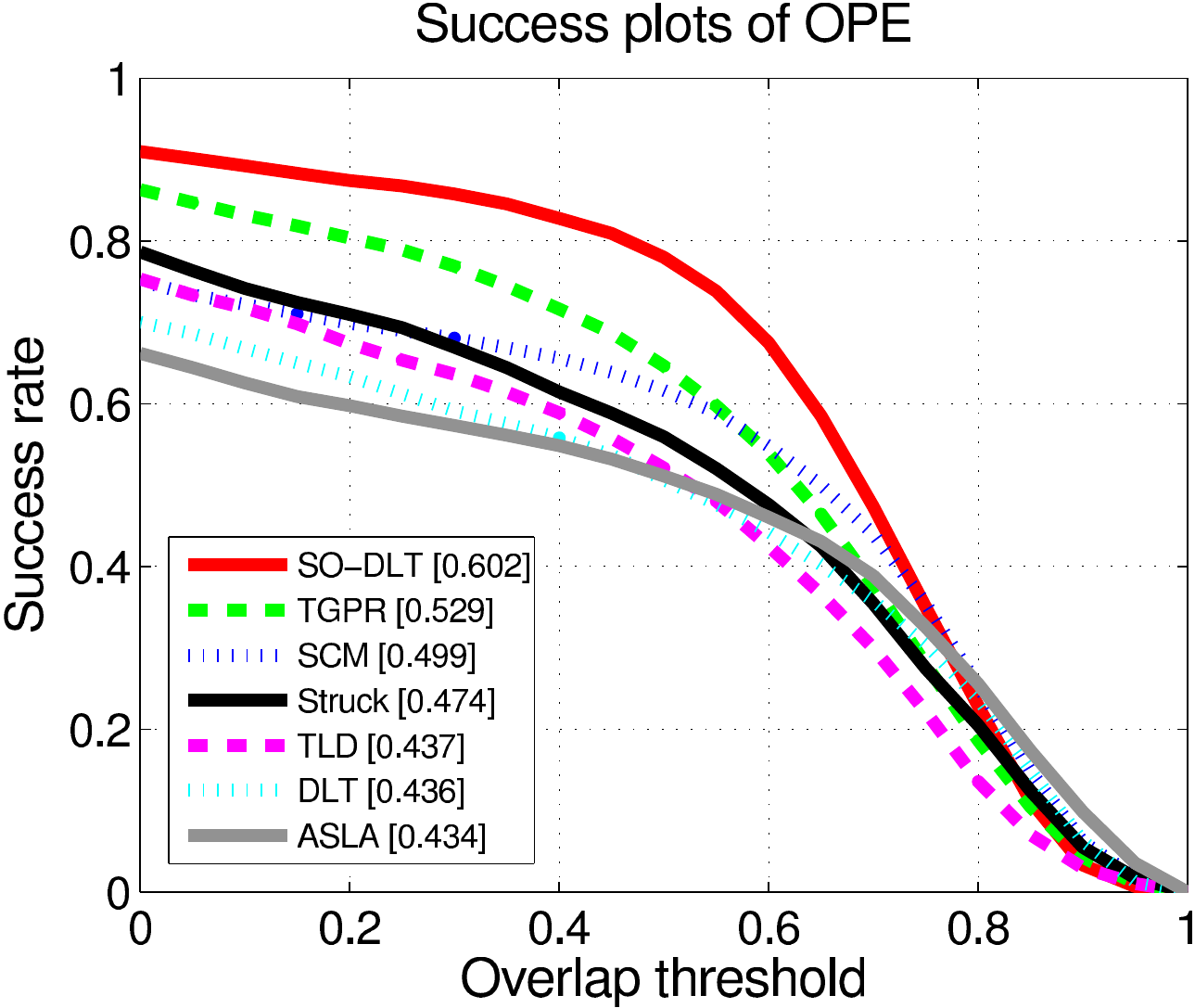}
	\includegraphics[width=0.47\linewidth]{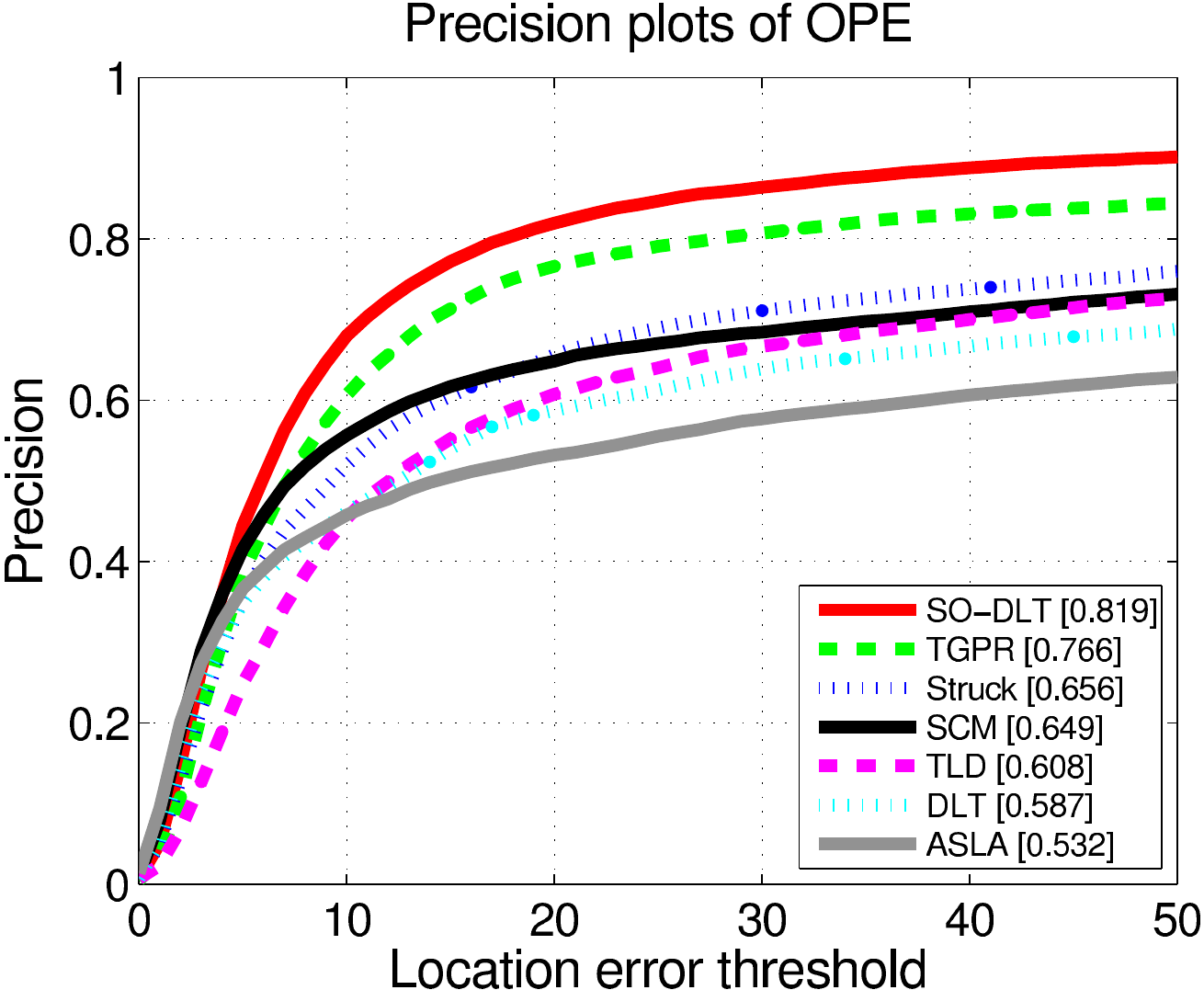}
	\caption{\label{fig:overall_OPE}Plots of OPE on the CVPR2013 Visual Tracker Benchmark. The performance  score for each tracker is shown in the legend. For success plots the score is the AUC value while for  precision plots the score is the precision value at threshold 20.  }
\end{figure}
\begin{figure}[htb]
	\centering
	\includegraphics[width = 0.8\linewidth]{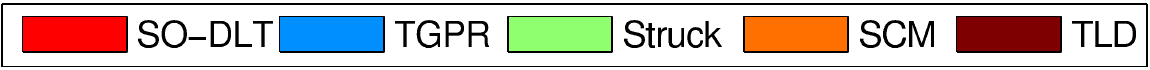}\\
	\subfigure[performance ranking scores based on success plots]{
		\includegraphics[width=0.47\linewidth]{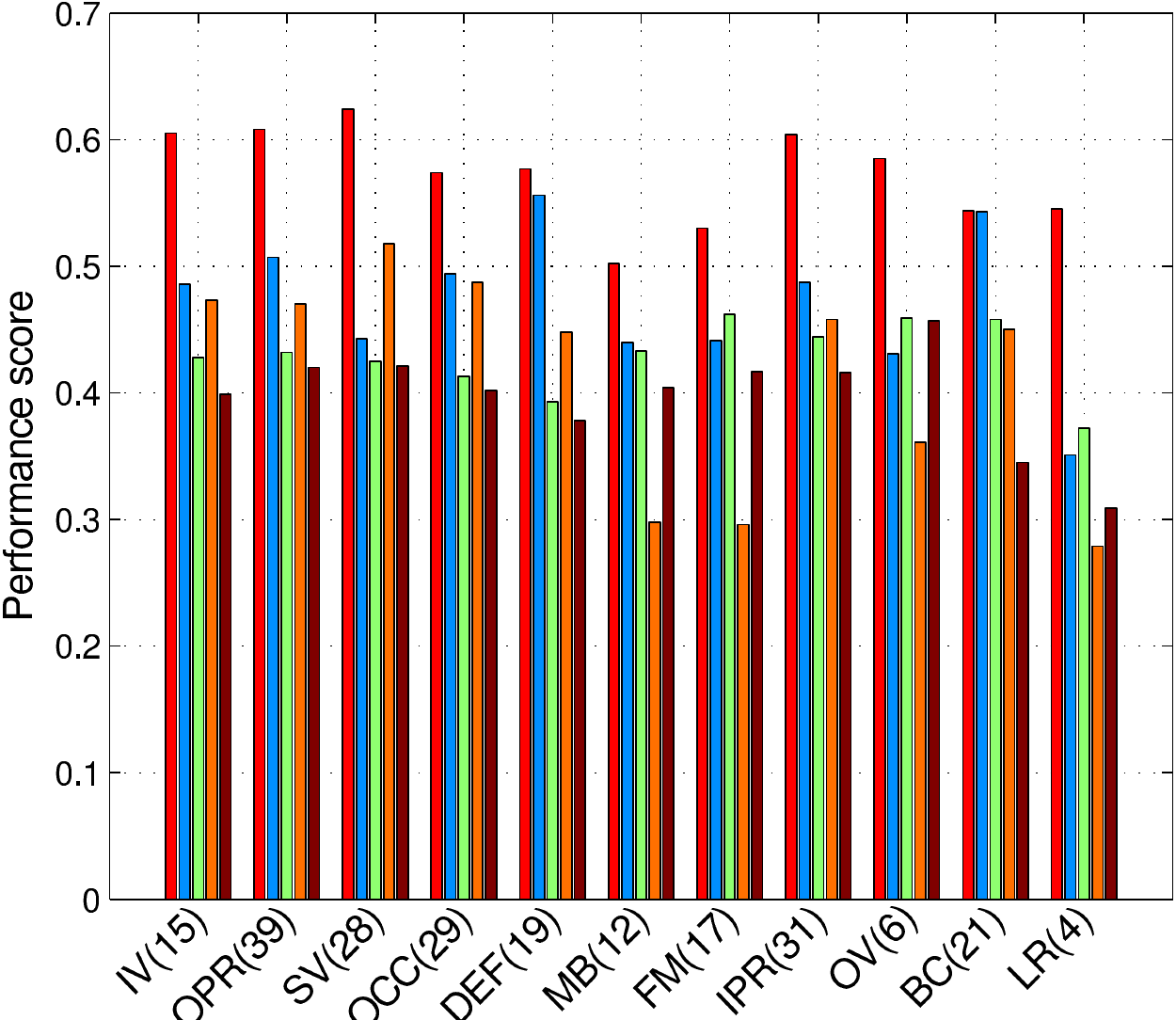}
	}
	\subfigure[performance ranking scores based on precision plots]{
		\includegraphics[width=0.47\linewidth]{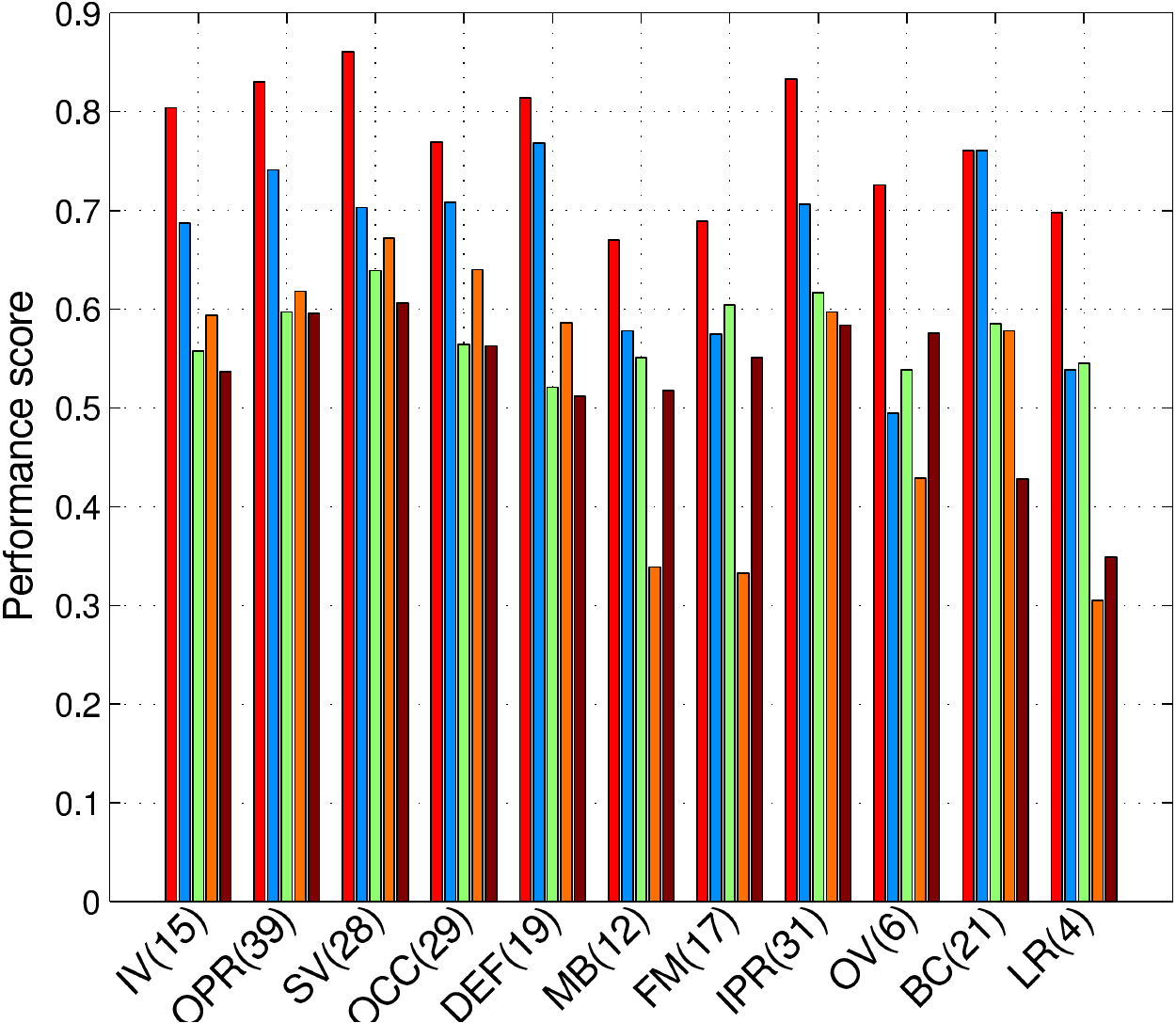}
	}
	\caption{\label{fig:attri_OPE}Average performance ranking scores of five leading trackers on different subsets of test sequences in OPE. Each subset of sequences corresponds to one of the attributes, namely, illumination variation (IV), out-of-plane rotation (OPR), scale variation (SV), occlusion (OCC), deformation (DEF), motion blur (MB), fast motion (FM), in-plane rotation (IPR), out-of-view (OV), background cluttered (BC), and low resolution (LR). The number after each attribute name is the number of sequences in the corresponding subset. The trackers included here are selected based on their overall performance ranking scores in OPE. }
\end{figure}
\subsubsection{Evaluation Setting and Metrics} We use two performance measures analogous to the \textit{area under curve} (AUC) measure for the \textit{receiver operating characteristic} (ROC) curve. Specifically, for a given overlap threshold in $[0, 1]$,  a tracker is considered successful in a frame if its overlap rate exceeds the threshold. The success rate for a video measures the percentage of successful frames over the entire video. By varying the threshold gradually from 0 to 1, it gives a plot of the success rate against the overlap threshold for each tracker. A similar performance measure called precision plot is defined for the central pixel error which measures the distance in pixels between the centers of the bounding boxes for the ground truth and the prediction. The difference is that the precision at threshold 20 is used for the performance score instead of the AUC score as in the success plots.
The results of 29 trackers on the benchmark can be found in~\cite{benchmark}. For a more complete comparison, we also include a recent tracker called TGPR~\cite{gpr}. As far as we know, TGPR is the best single tracker with code publicly available.

\subsubsection{Quantitative Results} Due to space limitations, we only show in each plot the overall performance of \emph{one pass error} (OPE) for our proposed tracker and some of the state-of-the-art trackers, which are TGPR~\cite{gpr} and the top 5 trackers using the CVPR2013 benchmark~\cite{benchmark}, namely, Struck~\cite{struck}, SCM~\cite{zhong2012robust}, TLD~\cite{tld}, ASLA~\cite{jia2012visual}, and DLT~\cite{dlt}.  The success and precision plots are shown in Fig.~\ref{fig:overall_OPE}.  For each tracker, a curve is obtained by averaging over those for all 50 test sequences.  From Fig.~\ref{fig:overall_OPE}, we can see that SO-DLT outperforms the other trackers by a large margin.  Specifically, SO-DLT outperforms the second best tracker TGPR by 13.8\% for the success plots and by 6.9\% for the precision plots.  As for the other top-ranking trackers in the benchmark, the improvement of SO-DLT is even more significant.

We also report in Tab.~\ref{tbl:overlapping} and Tab.~\ref{tbl:cpe} the average success rates at several thresholds for different methods.  SO-DLT consistently outperforms other trackers by a large margin.  Consider, for example, the case when the overlap rate is 0.5 which is a common choice for computing the success rate, SO-DLT outperforms the nearest baseline by 20.6\%.  We believe the comparison is substantial enough to demonstrate the superiority of SO-DLT.

\rowcolors{2}{white}{gray!25}
\begin{table}[tb]
	\begin{center}
	\begin{small}
	\begin{tabular}{lccc}
				&p@0.3	&p@0.5	&	p@0.7\\
		SO-DLT	&\textbf{0.8576}	&\textbf{0.7798}	&	\textbf{0.4732}\\
		TGPR		&0.7690	&0.6463	&	0.3767\\
		SCM		&0.6807	&0.6162	&	0.4396\\
		Struck		&0.6694	&0.5593	&	0.3543\\
		TLD		&0.6362	&0.5210	&	0.2997\\
		DLT		&0.5912	&0.5066	&	0.3581\\
		ASLA		&0.5730	&0.5112	&	0.3884\\
	\end{tabular}
	\end{small}
	\vspace{2mm}
	\caption{Success rates at different thresholds based on the overlap rate metric for different tracking methods.} \label{tbl:overlapping}
	\end{center}
\end{table}

\rowcolors{2}{white}{gray!25}
\begin{table}[tb]
	\begin{center}
	\begin{small}
	\begin{tabular}{lccc}
				&p@15	&p@25	&	p@35\\
				SO-DLT & \textbf{0.7721}	& \textbf{0.8470}	& \textbf{0.8772}\\
				TGPR & 0.7133	& 0.7907	& 0.8215\\
				Struck & 0.6047	& 0.6895	& 0.7257\\
				SCM & 0.6171	& 0.6703	& 0.6980\\
				TLD & 0.5520	& 0.6403	& 0.6848\\
				DLT & 0.5400	& 0.6128	& 0.6541\\
				ASLA & 0.5050	& 0.5554	& 0.5915\\
	\end{tabular}
	\end{small}
	\vspace{2mm}
	\caption{Success rates at different thresholds based on the central-pixel error metric for different tracking methods.} \label{tbl:cpe}
	\end{center}
\end{table}

For better analysis of the strengths and weaknesses of each tracker, each of the 50 sequences is also annotated with attributes that reflect the challenging factors. Fig.~\ref{fig:attri_OPE} shows the  ranking scores of the leading trackers on different groups of sequences, where each group corresponds to a different attribute. 
From Fig.~\ref{fig:attri_OPE}, we can see that SO-DLT substantially outperforms the other state-of-the-art trackers under almost all conditions. More precisely, our proposed tracker gets the highest score for all 11 attributes. Besides, we can also see that SO-DLT is especially good at handling some attribute groups such as ``illumination variation (IV)'', ``out-of-plane rotation (OPR)'', ``in-plane rotation (IPR)'', ``out-of-view (OV)'', and ``scale variation (SV)''.  The sequences in these groups often suffer from large appearance changes. This promising result demonstrates the great representational power of CNN in detecting the target object even for some extreme cases. Moreover, when model drifting happens, the proposed differentially-paced fine-tuned CNN can also help correct the drifting problem as soon as occlusion disappears.

Due to space limitations, we only include the representative results above and leave more details to the supplementary material.

\subsection{Non-rigid Object Tracking Dataset}
To gain a deeper insight into the proposed SO-DLT, we conduct an additional analysis on a challenging non-rigid object tracking dataset proposed in~\cite{hough}.

In this experiment, we only compare with TGPR~\cite{gpr}, which is the best tracker among all other trackers compared using the previous benchmark, and PixelTracker~\cite{pixeltrack}, which is a state-of-the-art non-rigid object tracker. Since the target is highly deformable, it is not meaningful to use the overlap rate defined by bounding boxes as a performance measure.  So we only use the central pixel error as our evaluation metric. The results are shown in Tab.~\ref{tbl:nonrigid}.

\rowcolors{2}{white}{gray!25}
\begin{table}[tb]
	\begin{center}
	\begin{small}
	\begin{tabular}{lccc}
					&SO-DLT	&TGPR	&PixelTracker	\\
		cliff-dive1		&\textbf{14.36}	&32.77	&	23.12\\
		cliff-dive2		&\textbf{15.04}	&17.65	&	35.86\\
		motocross1		&\textbf{14.49}	&188.09	&	125.71\\
		motocross2		&12.18	&\textbf{11.95}	&	31.76\\
		skiing			&\textbf{5.85}			&282.00	&	7.22\\
		mountain-bike	&6.10		&\textbf{5.48}		&	215.05\\
		volleyball		&83.77	&\textbf{7.24}		&	115.86\\
		\hline\hline
		average		&\textbf{21.69}		&77.88		&79.26
	\end{tabular}
	\end{small}
	\caption{Central pixel errors of different trackers on the non-rigid object tracking dataset.} \label{tbl:nonrigid}
	\end{center}
	\vspace{-4mm}
\end{table}

\begin{figure*}[htb]
	\centering
	{\footnotesize \rotatebox{90}{~~~~(a)dive1}}~\includegraphics[width=0.14\linewidth]{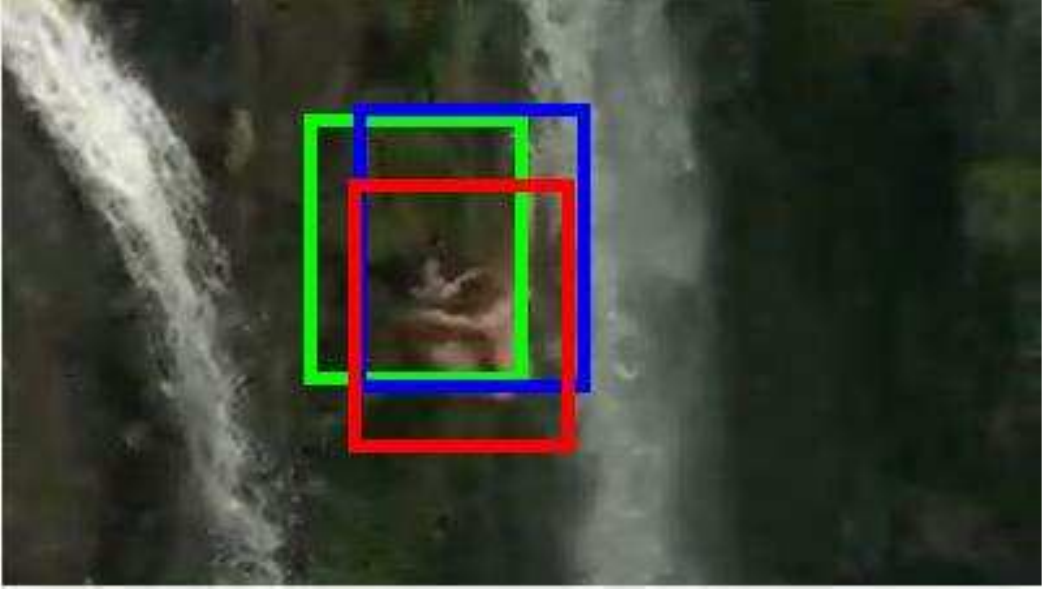}
	\includegraphics[width=0.14\linewidth]{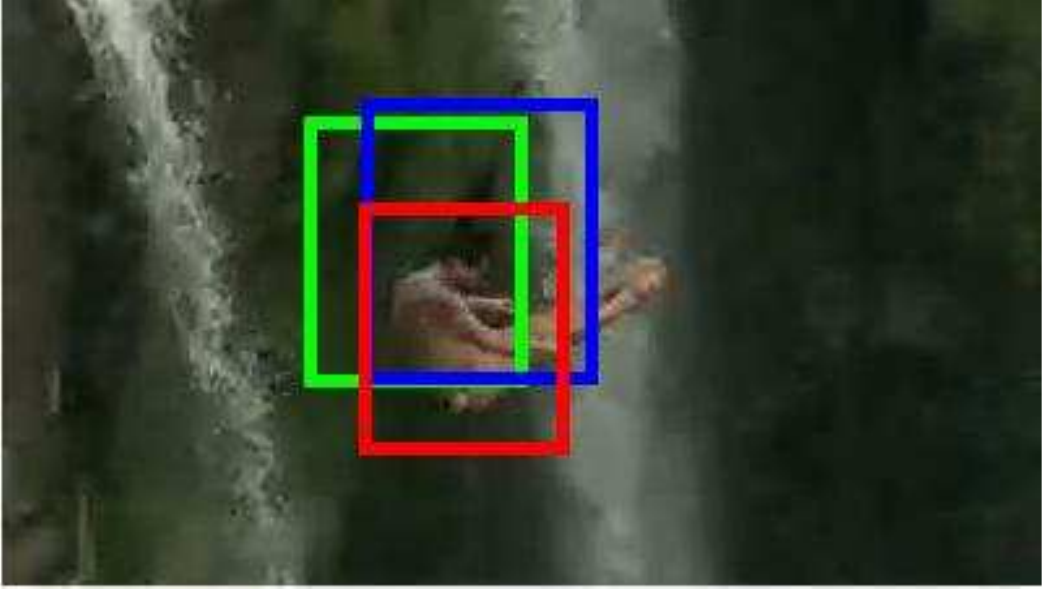}
	\includegraphics[width=0.14\linewidth]{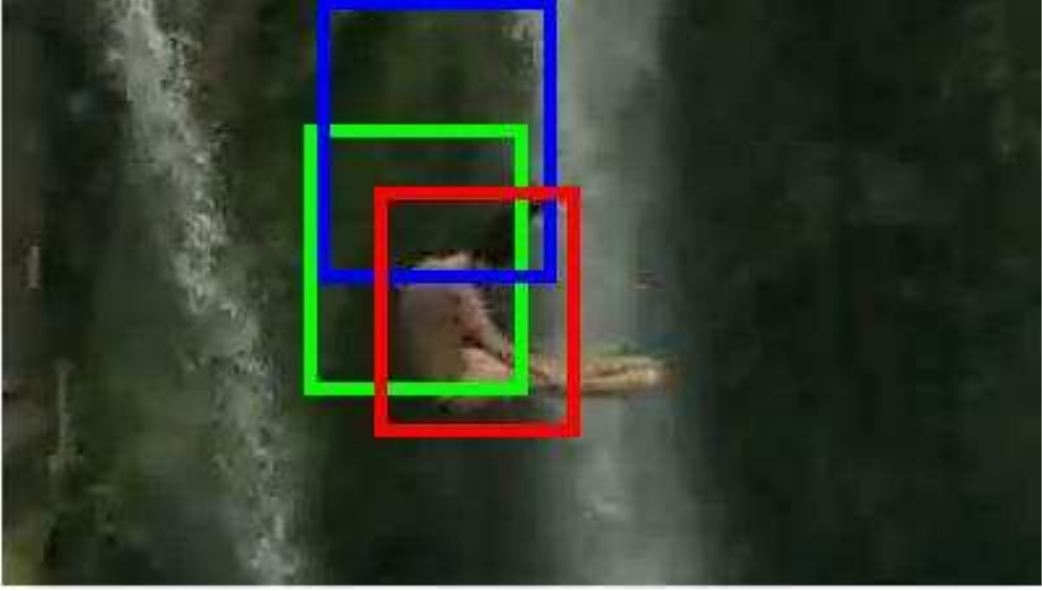} 
	\includegraphics[width=0.14\linewidth]{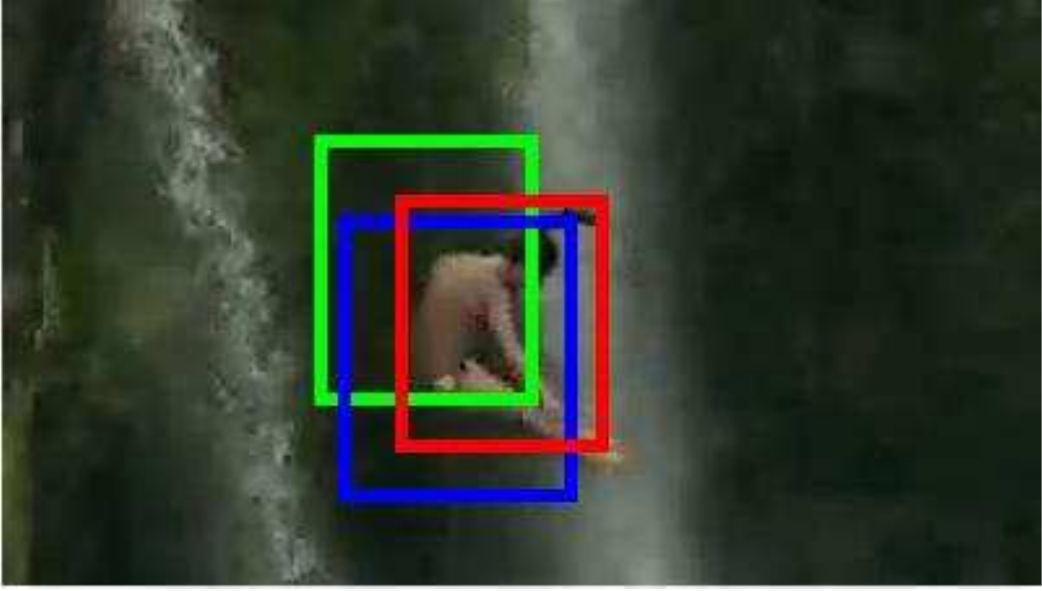} 
	\includegraphics[width=0.14\linewidth]{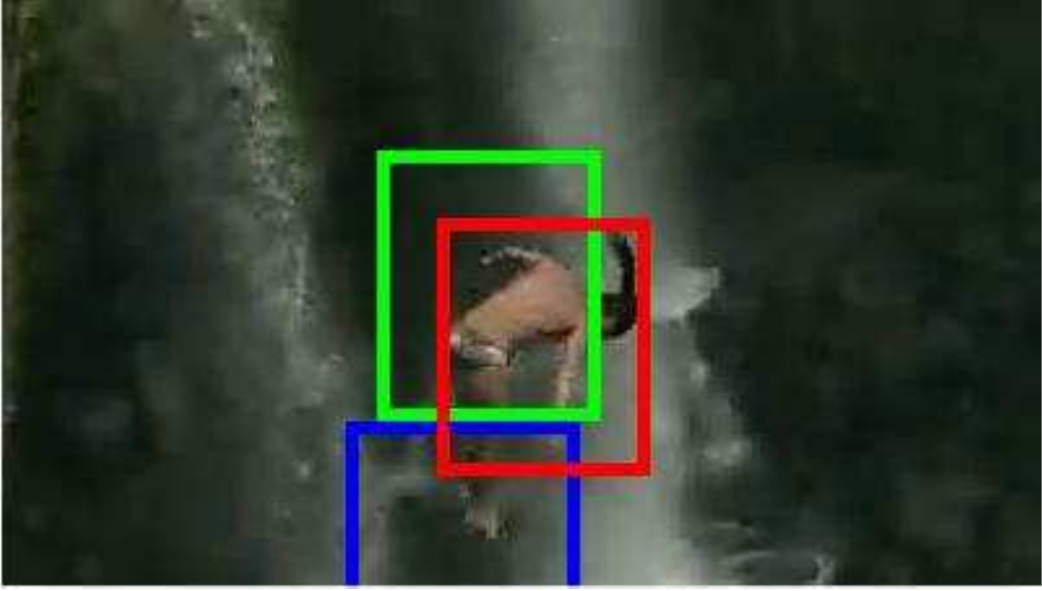}
	\includegraphics[width=0.14\linewidth]{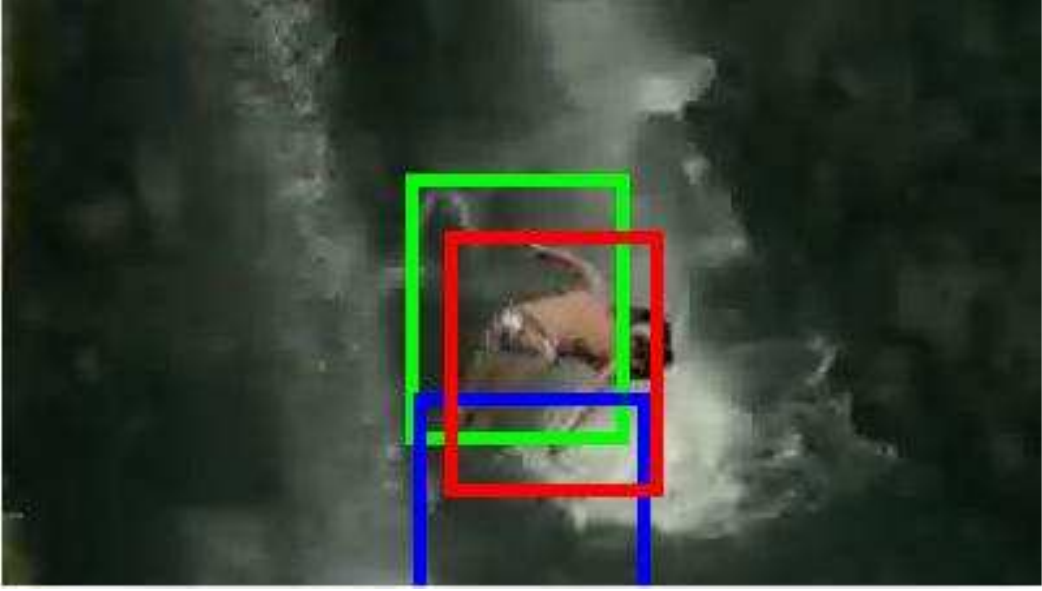} \\
	{\footnotesize \rotatebox{90}{~~~~(b)dive2}}~\includegraphics[width=0.14\linewidth]{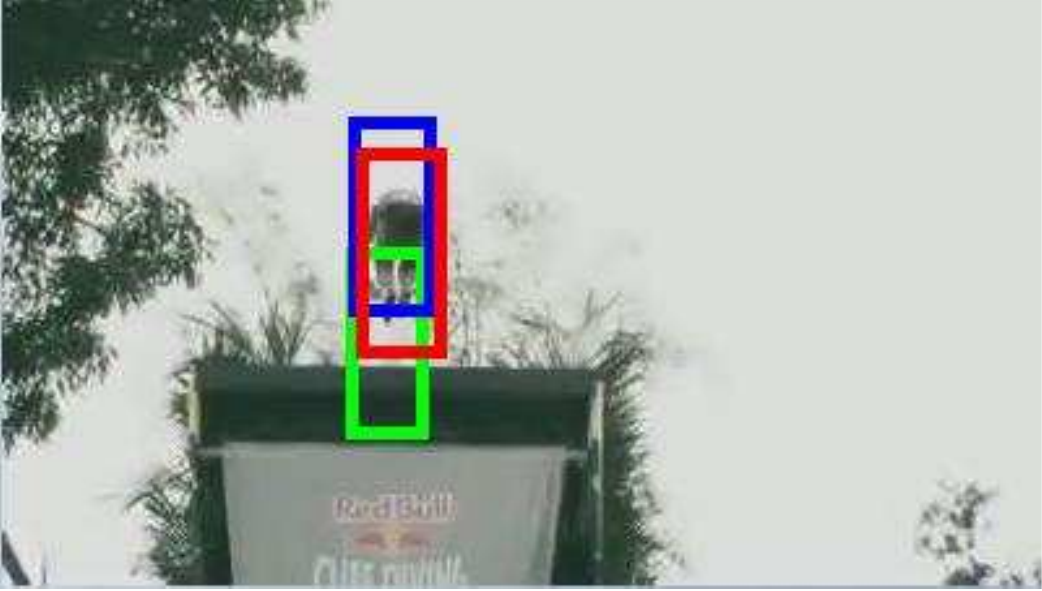}
	\includegraphics[width=0.14\linewidth]{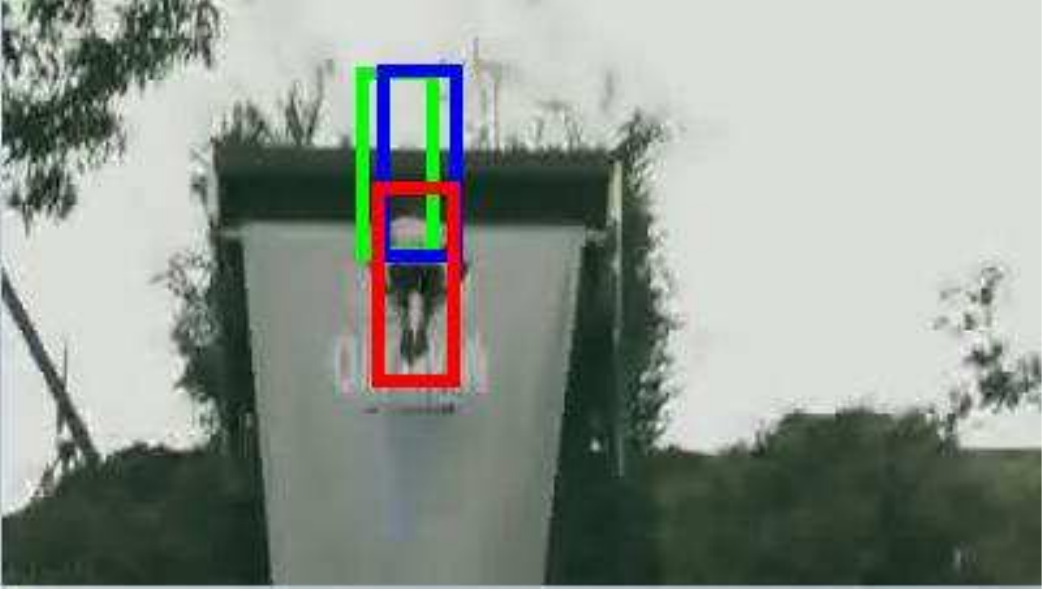}
	\includegraphics[width=0.14\linewidth]{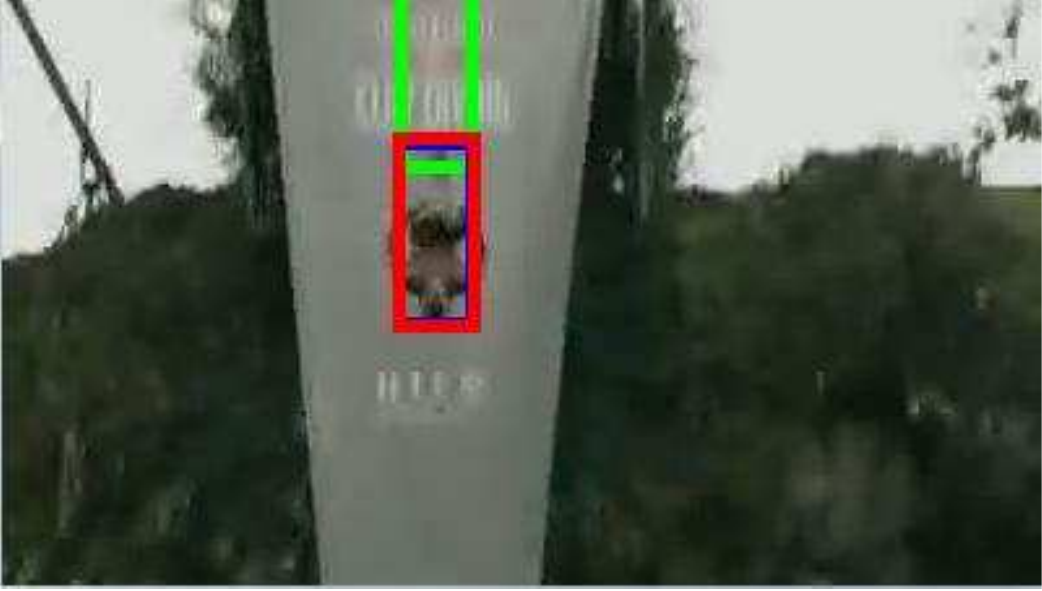} 
	\includegraphics[width=0.14\linewidth]{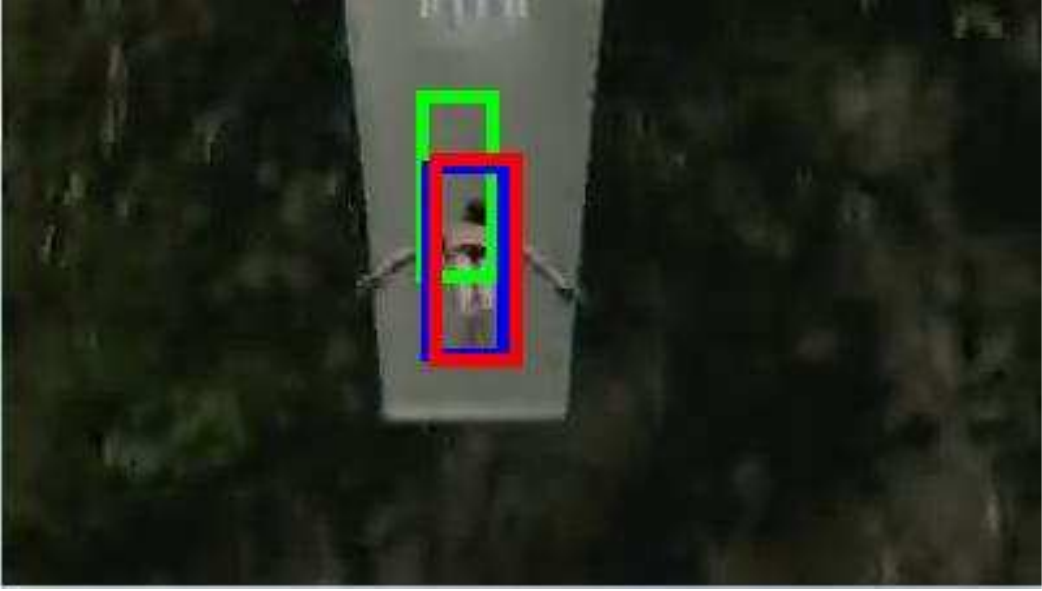} 
	\includegraphics[width=0.14\linewidth]{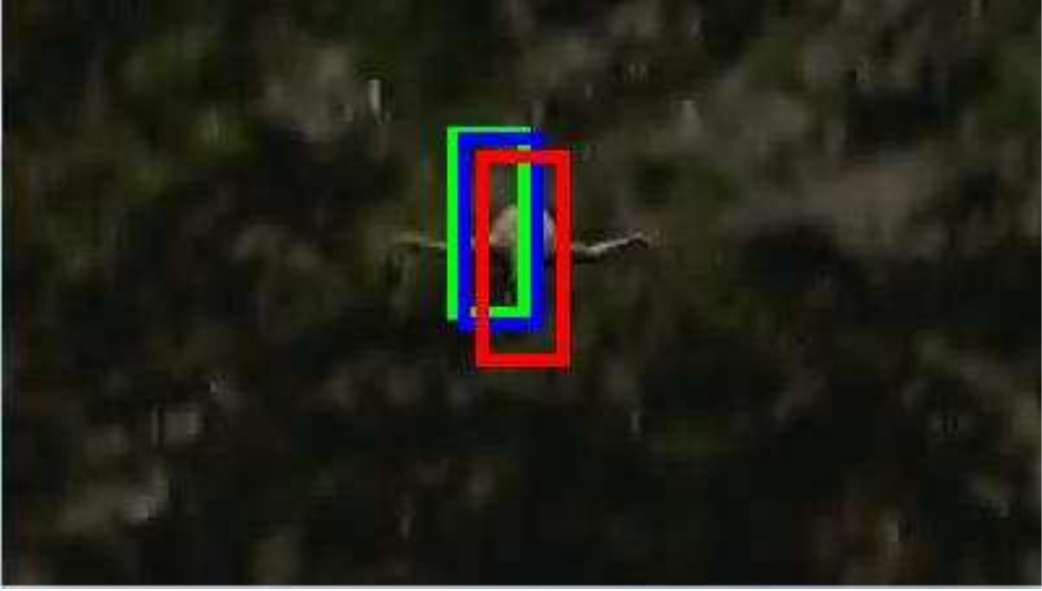}
	\includegraphics[width=0.14\linewidth]{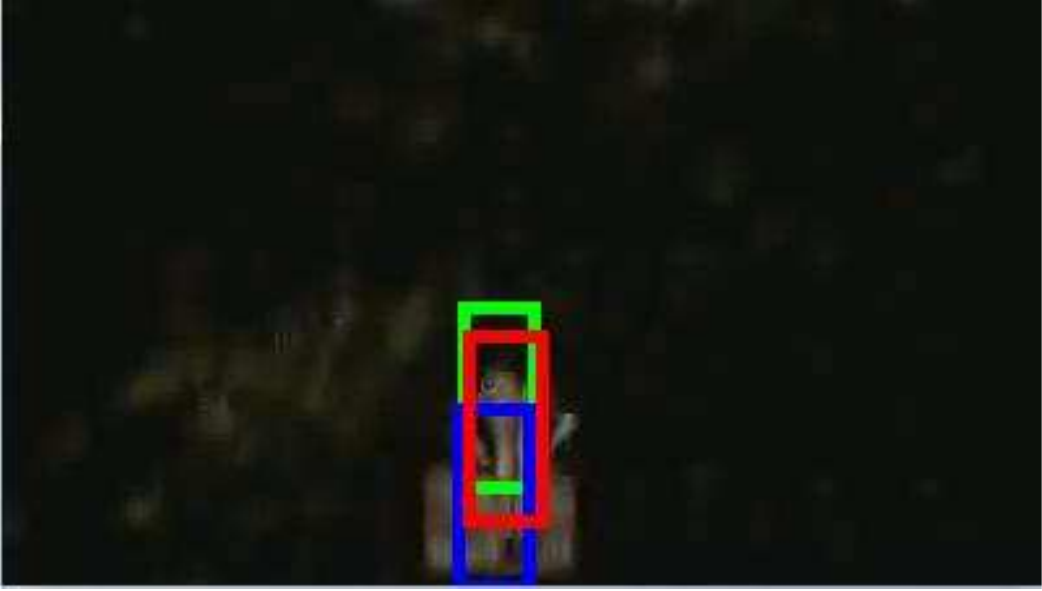} \\
	{\footnotesize \rotatebox{90}{~~~(c)motor2}}~\includegraphics[width=0.14\linewidth]{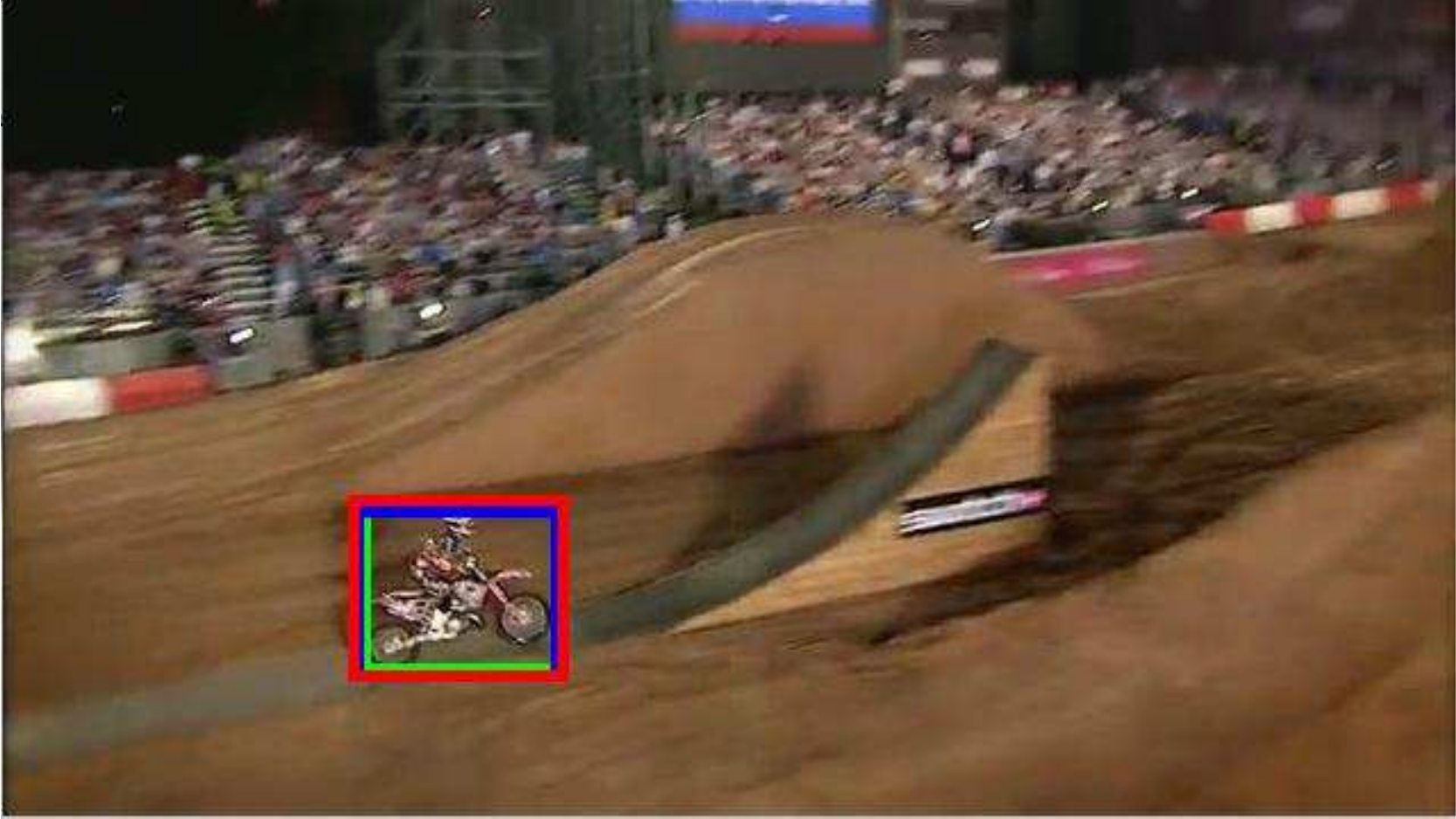}
	\includegraphics[width=0.14\linewidth]{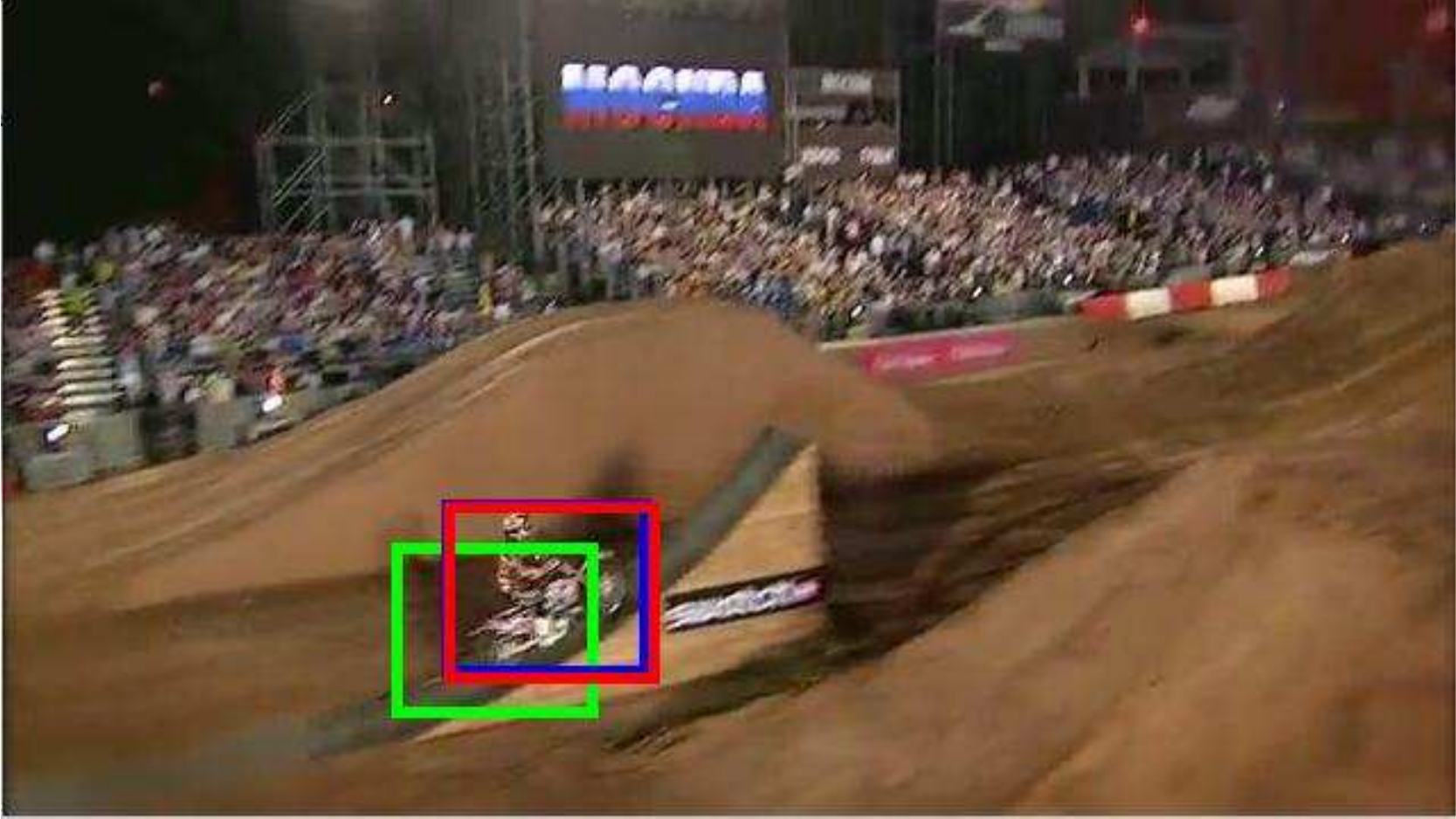}
	\includegraphics[width=0.14\linewidth]{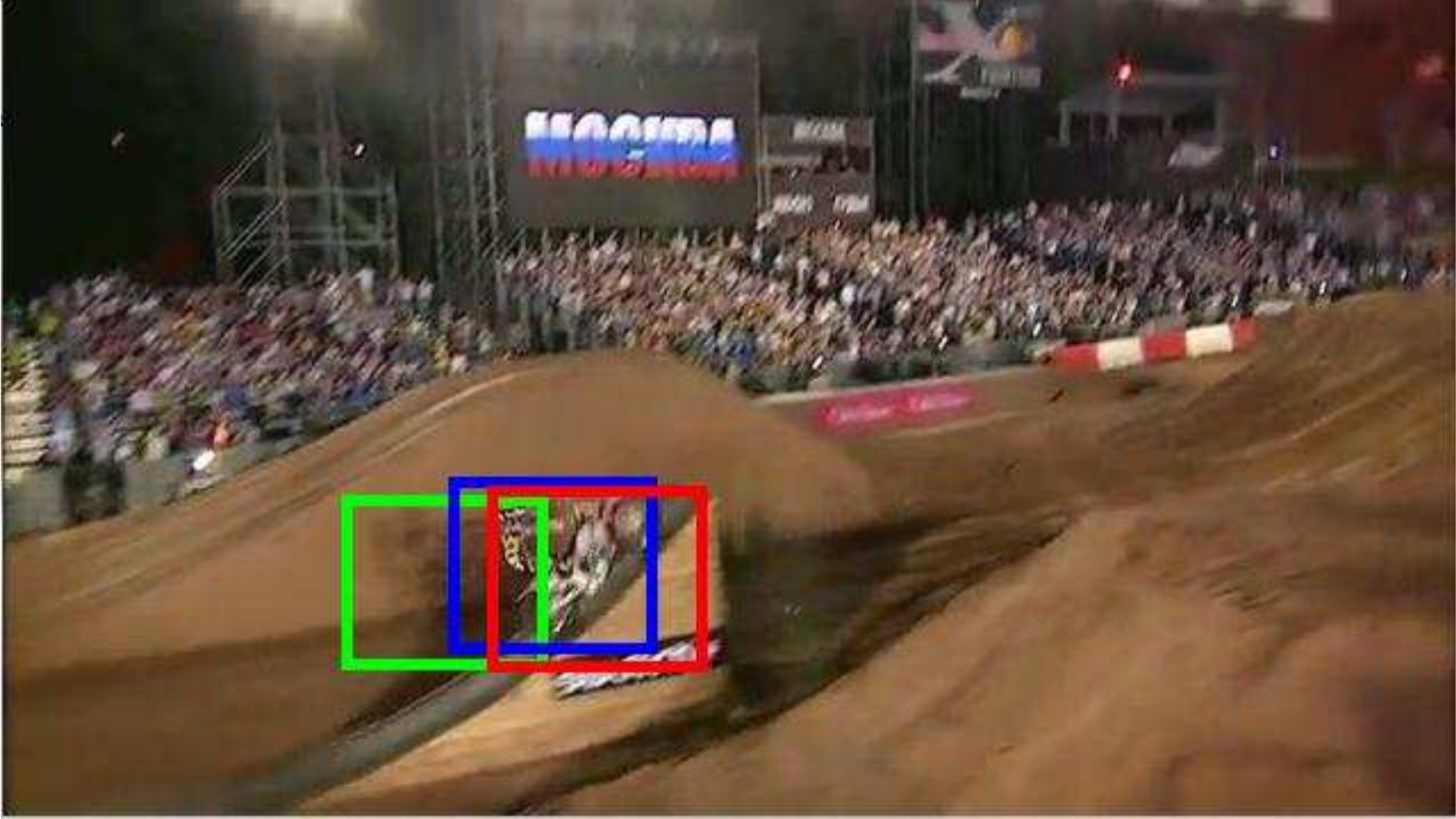} 
	\includegraphics[width=0.14\linewidth]{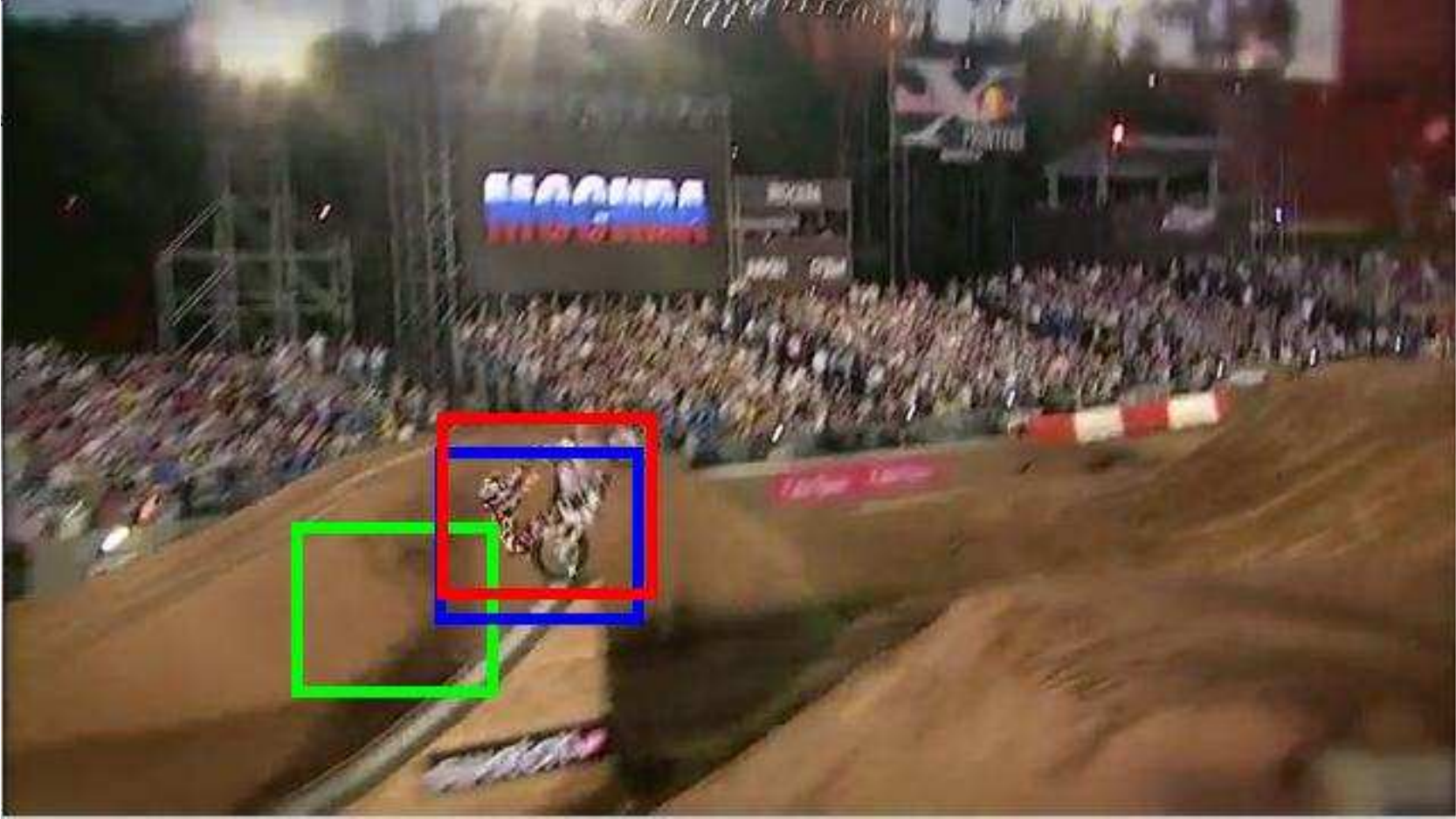} 
	\includegraphics[width=0.14\linewidth]{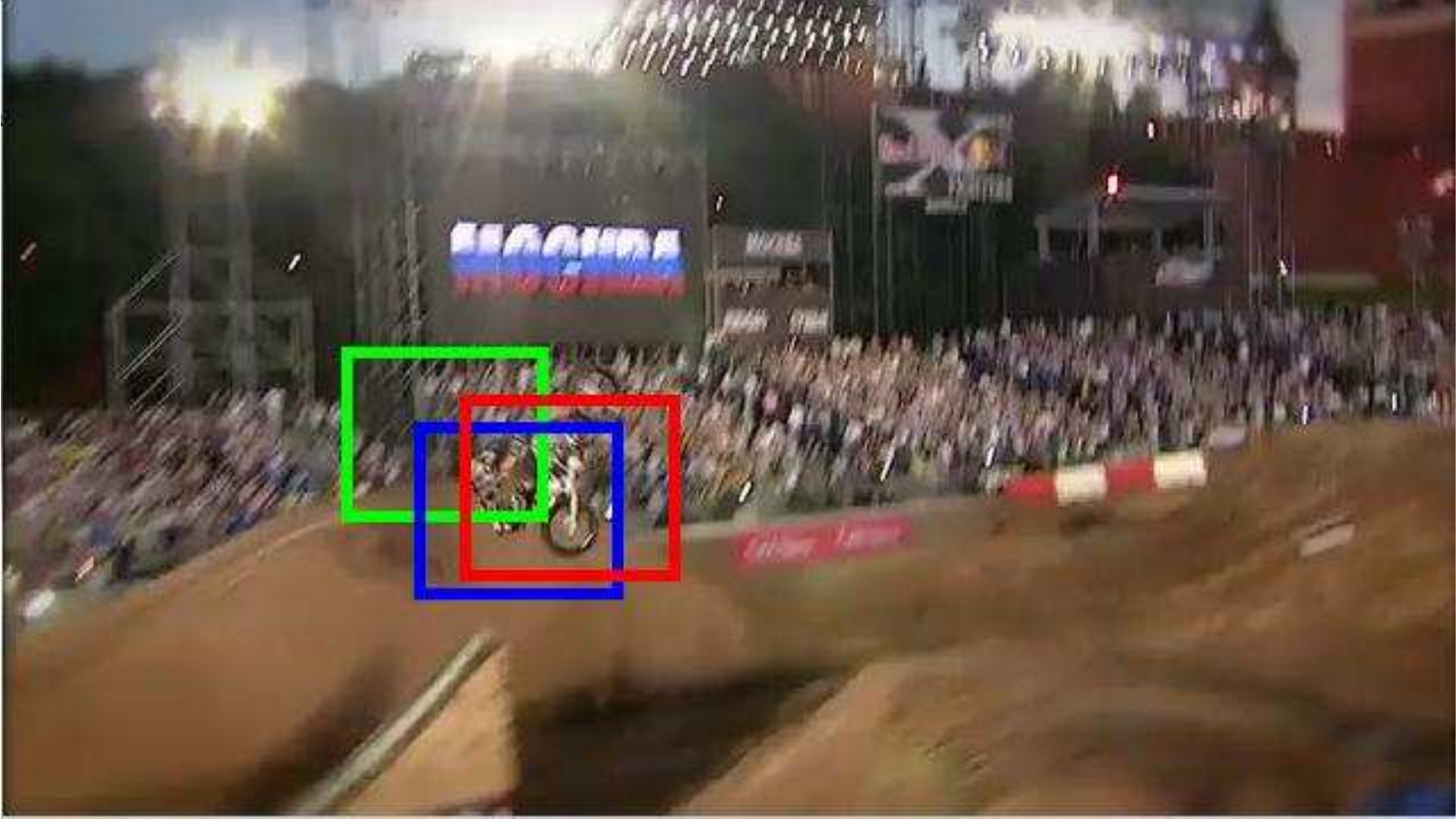}
	\includegraphics[width=0.14\linewidth]{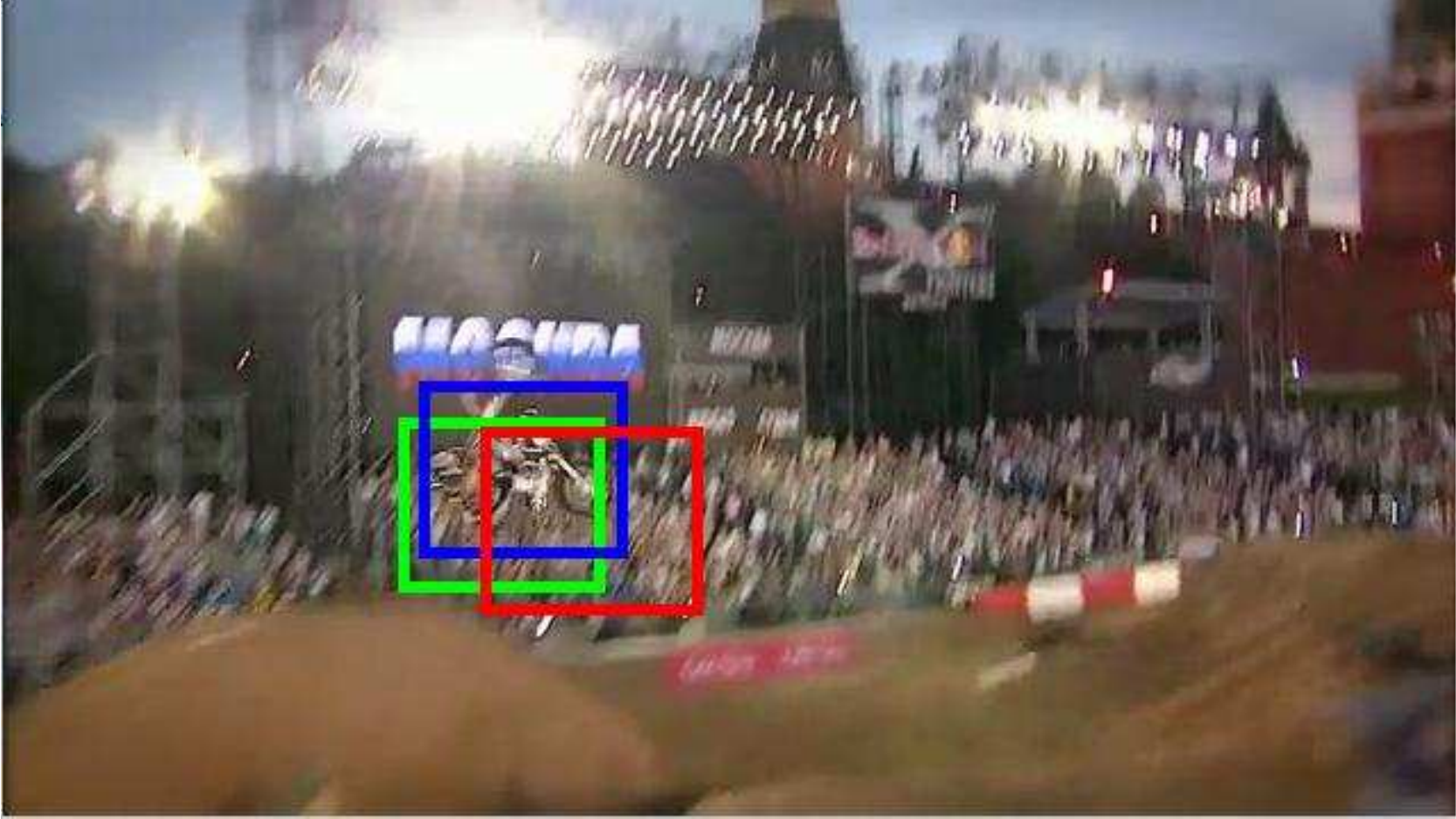}
	
	\vspace{2mm}
	\includegraphics[width=0.35\linewidth]{legend_video}\\	
	\caption{Results of some test sequences in the non-rigid tracking dataset.}
	\label{fig:nonrigid}
\end{figure*}

As we can see, although it is not specially designed for tracking non-rigid objects, SO-DLT can still outperform PixelTracker~\cite{pixeltrack}. Compared to the generic tracker TGPR~\cite{gpr}, the improvement is even more significant. SO-DLT can always track the target to the end with the exception of the \emph{volleyball} sequence. We show some visual results in Fig.~\ref{fig:nonrigid}.

\subsection{Discussion}
Although our tracker demonstrates superior performance on the two benchmarks, some failure cases do exist and a detailed analysis of them is important for making further improvement.  Here are some such cases:
\begin{enumerate}
	\item The tracker is likely to drift when there exist distractors in the background and when the target is occluded.  Due to the strong, invariant representational power of CNN, the distractors with similar appearance may be incorrectly recognized as the target, leading to incorrect tracking results.
	\item The tracker can possibly drift if the initial bounding box is not properly specified. This case often occurs when the shape of the target is quite irregular so that only a small portion of the object is specified by the bounding box as the target. 
\end{enumerate}
To address these challenges, here are some possible solutions: 
\begin{enumerate}
	\item In addition to directly transferring high-level features in SO-DLT which may introduce too much invariance to cases that contain distractors, we may build another tracker to transfer only low-level features that capture the local invariance of object appearance.  The two trackers can then work collaboratively in making the final decision. 
	\item More advanced model update methods may be tried. For example, model ensemble based methods~\cite{meem, ebt} have shown great potential recently. They can detect and correct the mistakes made during model update by undoing some incorrect model updates or re-weighting each model adaptively.
	\item To better handle irregular object shapes, our model may output a pixel-wise map of the target in case appropriate data from the first frame is available for pre-training and initialization.
\end{enumerate}
We will investigate these and possibly other ideas in our future work.

\section{Conclusion}

In this paper, we have exploited the effectiveness of transferring high-level feature hierarchies for visual tracking. To the best of our knowledge, we are the first to bring large-scale CNN to the area of visual tracking and show significant improvement over the state-of-the-art trackers. Instead of modeling tracking as a proposal classification problem, we presented a novel structured output CNN for visual tracking. Moreover, instead of learning to reconstruct the input image as in previous work, the CNN is first pre-trained on the large-scale ImageNet detection dataset~\cite{imagenet} to learn to localize objects so as to alleviate the problem caused by lack of labeled training data.
This objectness CNN is then transferred and fine-tuned during the online tracking process. Extensive experiments have validated the superiority of our SO-DLT tracker. 
{\small
\bibliographystyle{ieee}
\bibliography{sodlt}
}

\end{document}